\documentclass[10pt,twocolumn,letterpaper]{article}

\usepackage[final]{cvpr}              

\usepackage{graphicx}
\usepackage{amsmath}
\usepackage{amssymb}
\usepackage{booktabs}
\usepackage{amsmath}
\usepackage{amssymb}
\usepackage{comment}
\usepackage{bm}
\usepackage{bbm}
\usepackage{algorithm}
\usepackage{algpseudocode}
\usepackage{mathtools}
\usepackage{multirow}
\usepackage{multicol}
\usepackage{multirow}
\usepackage{pifont}

\newcommand{\cmark}{\text{\ding{51}}}%
\newcommand{\xmark}{\text{\ding{55}}}%

\newcommand{\gxmark}{\textcolor{gray}{\xmark}}
\usepackage[accsupp]{axessibility}  

\definecolor{cvprblue}{rgb}{0.21,0.49,0.74}
\usepackage[pagebackref,breaklinks,colorlinks,allcolors=cvprblue]{hyperref}

\newcommand\blfootnote[1]{%
  \begingroup
  \renewcommand\thefootnote{}\footnote{#1}%
  \addtocounter{footnote}{-1}%
  \endgroup
}

\def\paperID{2162} 
\def\confName{CVPR}
\def\confYear{2025}

\title{SmartEraser: Remove Anything from Images using Masked-Region Guidance}

\author{Longtao Jiang$^{1,*}$
\hspace{.1cm} Zhendong Wang$^{1,*}$
\hspace{.1cm} Jianmin Bao$^{2,*\dagger\heartsuit}$
\\ Wengang Zhou$^{1,\dagger}$
\hspace{.1cm} Dongdong Chen$^{2}$
\hspace{.1cm} Lei Shi$^{2}$
\hspace{.1cm} Dong Chen$^{2}$ 
\hspace{.1cm} Houqiang Li$^{1}$ \\
\textsuperscript{1}University of Science and Technology of China
\hspace{.1cm} \textsuperscript{2}Microsoft Research Asia
\\ {\tt\small \url{https://longtaojiang.github.io/smarteraser.github.io/}}
\vspace{-2pt}
}

\begin{document}

\twocolumn[{%
\renewcommand\twocolumn[1][]{#1}%
\maketitle
\begin{center}
  \centering
  \vspace{-1.5em}
  \captionsetup{type=figure}
   \includegraphics[width=1\linewidth]{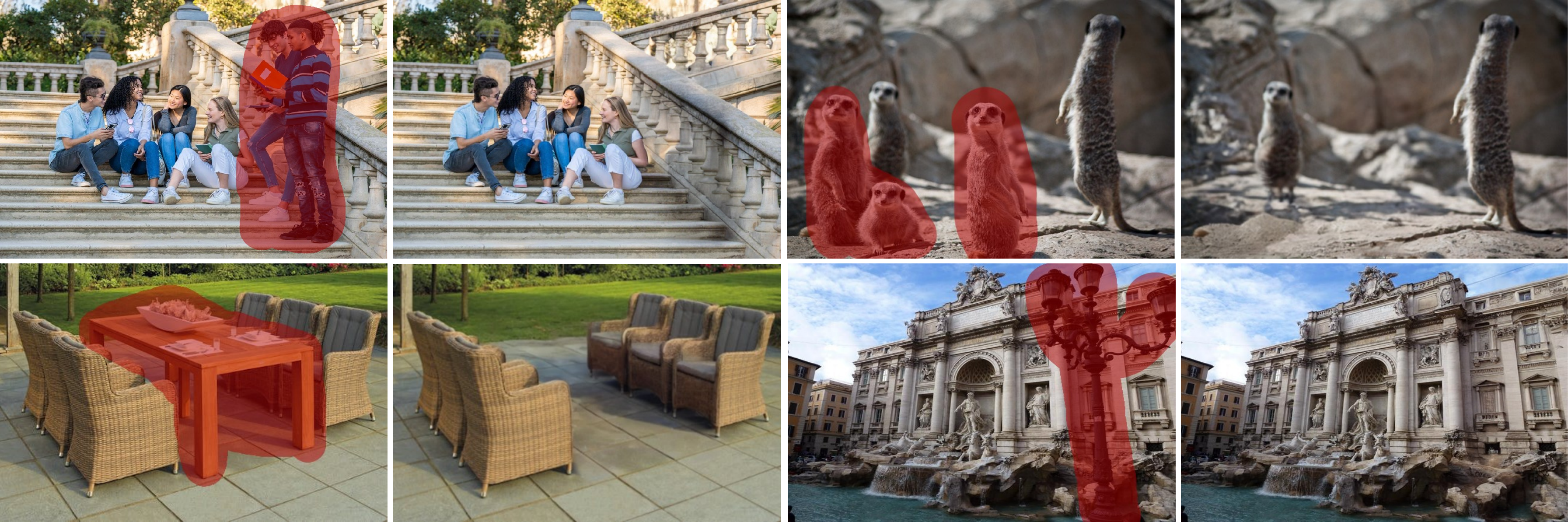}
   \vspace{-2em}
   \caption{Object removal results generated by SmartEraser using user-provided masks. SmartEraser not only generates photorealistic images with the target objects removed but also better preserves the surrounding context around the removed objects.}
   \label{fig:teasor}
\end{center}
}]

\begin{abstract}
Object removal has so far been dominated by the ``mask-and-inpaint" paradigm, where the masked region is excluded from the input, leaving models relying on unmasked areas to inpaint the missing region. However, this approach lacks contextual information for the masked area, often resulting in unstable performance. In this work, we introduce SmartEraser, built with a new ``\textbf{removing}" paradigm called Masked-Region Guidance. This paradigm retains the masked region in the input, using it as guidance for the removal process. It offers several distinct advantages: (a) it guides the model to accurately identify the object to be removed, preventing its regeneration in the output; (b) since the user mask often extends beyond the object itself, it aids in preserving the surrounding context in the final result. Leveraging this new paradigm, we present Syn4Removal, a large-scale object removal dataset, where instance segmentation data is used to copy and paste objects onto images as removal targets, with the original images serving as ground truths.
Experimental results demonstrate that SmartEraser significantly outperforms existing methods, achieving superior performance in object removal, especially in complex scenes with intricate compositions.
\end{abstract}

\section{Introduction}
\label{sec:intro}
With the rapid advancement of generative models~\cite{latentdm,vqgan,vqvae,gan,ddpm,song2023consistency,ramesh2022hierarchical,sdxl}, image editing has garnered significant attention for its broad applications. Among various editing tasks, object removal~\cite{powerpaint,clipaway} is especially prominent, empowering users to seamlessly eliminate unwanted elements while maintaining the realism of original images. This functionality, essential for removing distracting objects in photos, has become a key feature in widely used applications like Photoshop, Google Photos, Microsoft Designer.
\vspace{-2em}

\blfootnote{$^*$ Equal contribution.\ \ $\dagger$ Corresponding authors.\ \ $\heartsuit$ Project leader.}

\begin{figure}[t]
  \centering
   \includegraphics[width=1\linewidth]{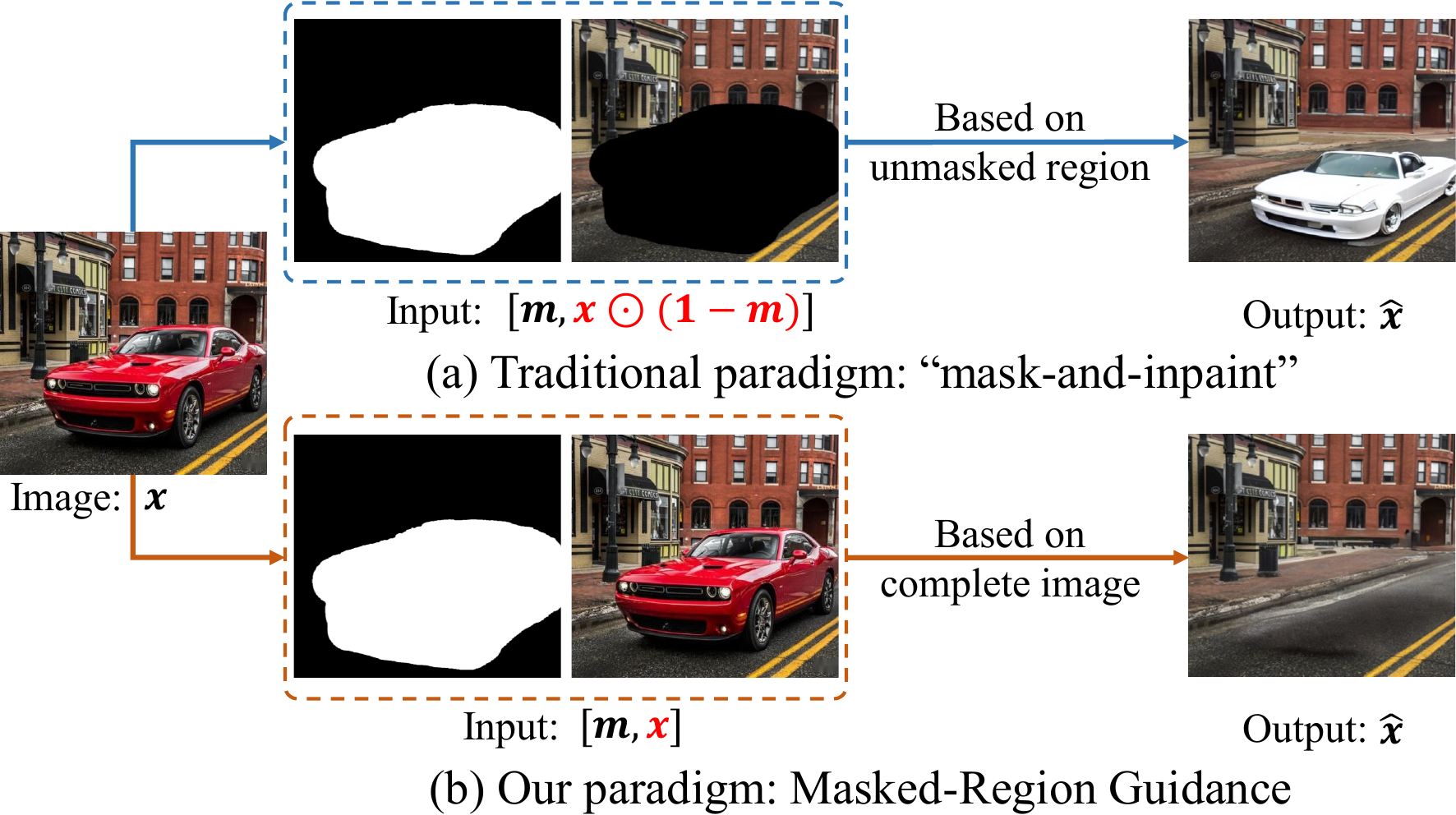}
   \vspace{-2em}
   \caption{Comparison of the ``\emph{mask-and-inpaint}" paradigm and the proposed \emph{Masked-Region Guidance} paradigm for object removal. Unlike the \emph{mask-and-inpaint} approach, our paradigm uses the masked regions as critical guidance for object removal.}
   \vspace{-2em}
   \label{fig:introduction}
\end{figure}

Currently, most object removal methods~\cite{lama,bld,repaint,powerpaint,clipaway} adopt a “mask-and-inpaint” paradigm, where the masked region is excluded from the input and typically filled with a neutral placeholder (\emph{e.g.}, black). The model then inpaints the masked region based on the surrounding content. However, we identify two primary issues with this approach. \emph{First}, this strategy often generates unintended objects within the masked region. Lacking precise discrimination between removal targets and other content, these methods rely heavily on background context, sometimes leading to the unintentional inpainting of new objects—such as adding a new car on the road rather than removing the original one, as illustrated in Figure~\ref{fig:introduction}(a). \emph{Second}, user-defined masks frequently exceed the target object, requiring the ``mask-and-inpaint'' approach to synthesize these extended region. This will inadvertently modify nearby context, reducing visual coherence.

To address these limitations, we introduce a novel paradigm for object removal called \textit{Masked-Region Guidance}. The core idea is that the masked region should not be excluded but rather utilized as critical guidance during the removal process. Our method is straightforward: instead of replacing the masked region with a placeholder, we retain the original image as input, with the masked region indicated by a mask input, as in existing methods. As illustrated in Figure~\ref{fig:introduction}(b), this paradigm enables the model to accurately identify the target object, preventing unintended regeneration in the output and effectively preserving the surrounding context of the target object in the final result.

However, our new paradigm cannot be directly implemented with existing object removal data construction methods. Typically, their training data are created by masking parts of the image, and the model is trained to predict the masked content. If applying this approach to our proposed \textit{Masked-Region Guidance} paradigm, the model could exploit a shortcut by simply replicating the masked content from the input, due to the masked region being included in the input. Thus, a dataset consisting of triplets of an input, a mask, and the removal result is essential for our paradigm. Unfortunately, existing related datasets~\cite{rord,objectdrop,defacto,instinpainting} either contain a limited number of unique scenes or rely on inpainting models to generate pseudo-removed results. 

To solve this, we introduce a synthetic technique to create training data specifically for object removal. Our approach involves pasting object instances from various images onto different background images to form the input images, with the pasted instance masks serving as the input masks and the original background images designated as ground truths. Using this method, we generate Syn4Removal, a large-scale dataset comprising triplets of real background images, masks, and backgrounds with pasted objects. The design of Syn4Removal provides diverse scenes and supports effective training under our new paradigm, encouraging the model to accurately learn object removal without shortcuts.

To make Syn4Removal suitable for training object removal models, we design a pipeline to generate high-quality data. First, we filter out low-quality instances and background images. Then, we develop a method to calculate feasible pasting locations on the image, ensuring that objects do not overlap with instances in the pasted area, which helps prevent the model from regenerating unwanted objects. Last, the instance is pasted onto a background image with a blending algorithm. The resulting dataset consists of 1 million image triplets.

Using the Masked-Region Guidance paradigm and the Syn4Removal dataset, we design a framework based on text-to-image diffusion models for object removal. To enhance the model's robustness to varying mask shapes from user input, we introduce a mask enhancement technique that simulates different mask shapes over removal targets. Additionally, we incorporate CLIP-based visual guidance to assist the model in better understanding the removal targets. The resulting model called \textit{SmartEraser} outperforms previous methods significantly in both quantitative and qualitative evaluations. We choose the name \textit{SmartEraser} because the model can ``smartly" identify removal targets and remove it while preserving its surrounding region.

Our contributions are mainly summarized as follows:
\begin{itemize}
\item We propose Masked-Region Guidance, a novel paradigm for object removal, which effectively addresses issues of object regeneration and surrounding region distortion. 
\item We introduce Syn4Removal, a large-scale high-quality dataset for object removal, which contains over a million pairs of images across diverse scenes and object types.
\item We carefully design a framework based on the text-to-image diffusion model by introducing mask enhancement and clip-based visual guidance.
\item Extensive experiments demonstrate that SmartEraser significantly outperforms previous object removal methods in terms of quality and robustness.
\end{itemize}


\section{Related Work}
\label{sec:related_work}
\noindent\textbf{Image inpainting.}
Image inpainting aims to seamlessly fill missing regions in images by leveraging contextual information around the masked region. Based on the ``mask-and-inpaint'' paradigm, earlier works are mainly based on generative adversarial networks (GANs)~\cite{gan,cgan}. Typically, these methods~\cite{lama,mat,zits,patchmatch,yu2018generative,li2022misf,Liu_2022_CVPR,Dong_2022_CVPR,Wang_2022_CVPR} randomly mask part areas of images~\cite{zhou2017places,liu2015celeba,karras2017celebahq,karras2019ffhq} and then predict these masked regions. However, GAN-based approaches~\cite{Yu_2021_ICCV,Guo_2021_ICCV,Wan_2021_ICCV,Wang_2021_ICCV,Zeng_2021_ICCV} often suffer from limited diversity and quality in generated content, resulting in blurry or unrealistic inpainted results.

Recently, approaches~\cite{bld,latentdm,clipaway,powerpaint,repaint,ju2024brushnet,deJorge_2024_ECCV,Chen2024CATiffusion} based on diffusion models have attracted significant attention due to their superior capability in generating image details. 
Among them, Repaint~\cite{repaint} injects the known region into intermediate images at all timesteps throughout the diffusion generation. 
SD-Inpaint~\cite{latentdm} utilizes the mask and cropped image as input and fills the masked region through diffusion.
While these approaches produce detailed textures, the generated content within the masked region often remains uncertain, with new objects occasionally appearing.

\noindent\textbf{Object removal.}
The object removal task aims to eliminate specified objects from images based on user-provided masks, seamlessly repainting the removed area with unmasked content. 
For example, PowerPaint~\cite{powerpaint} uses learnable task embeddings from context-aware inpainting as the positive prompt and those from text-guided inpainting as the negative prompt to guide object removal. A recent work CLIPAway~\cite{clipaway} leverages AlphaCLIP~\cite{alphaclip} and IP-Adapter~\cite{ipadapter} to inject background information into the model via cross-attention, ensuring removal results coherent with the background context. The common characteristic of these methods is that they retain the traditional ``mask-and-inpaint'' paradigm by discarding the masked regions of input images. In this paper, we claim that this paradigm is not suitable for the object removal task exactly due to the missing context of the masked region, causing incomplete removal and inconsistent context in removal results. 

We also observe that some instruction-based methods~\cite{instinpainting, instructpix2pix} attempt to remove objects using text prompts instead of the masked region. While they use the original image as input, their models are trained with removal results generated by inpainting models, which limits their performance. We believe that our synthetic strategy for building removal datasets could enhance the performance of these models.

There are several works~\cite{instinpainting,defacto,rord,objectdrop} aiming to construct datasets for object removal. For instance, GQA-Inpaint~\cite{instinpainting} and DEFACTO~\cite{defacto} synthesize ground truths with the help of image inpainting models. However, we find that relying on image inpainting leads to the performance of object removal being highly limited by the effect of inpainting. Another two works RORD~\cite{rord} and ObjectDrop~\cite{objectdrop} acquire real-world triplets through photography, but the high collection costs result in fewer than 3.5$k$ unique scenes. To incorporate with the Masked-Region Guidance, we construct a large-scale, high-quality dataset called Syn4Removal through a set of carefully designed strategies, which includes about 1 million triplets featuring diverse scenes and authentic ground-truth backgrounds with carefully filtering and pasting strategies.

\begin{figure*}[t]
  \centering
   \includegraphics[width=1\linewidth]{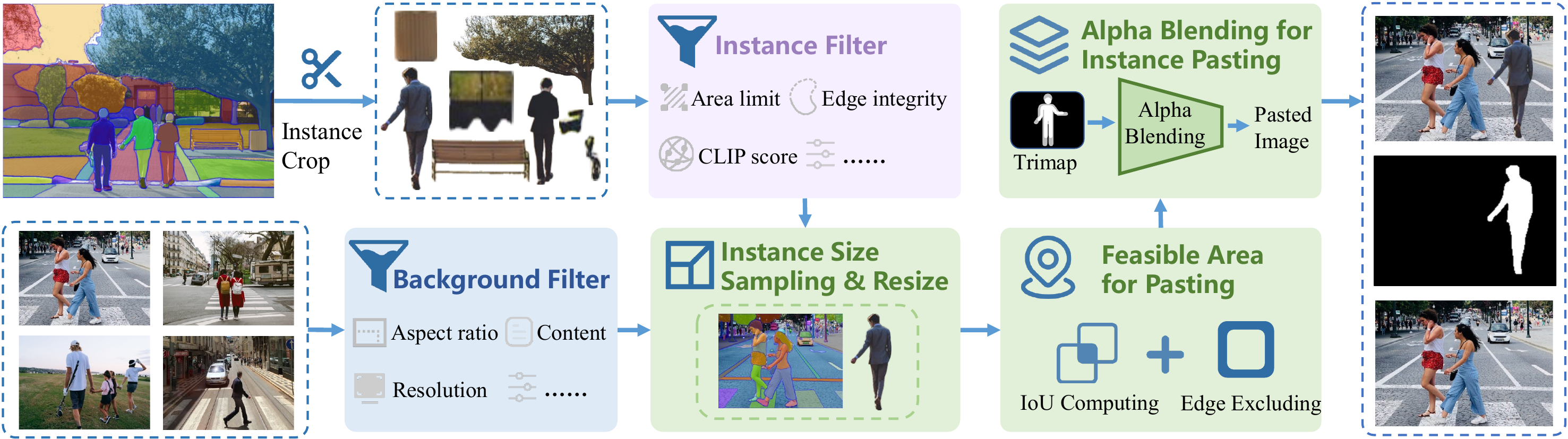}
   \vspace{-2em}
   \caption{The data generation pipeline of Syn4Removal. We apply instances and images to construct a triplet consisting of an input image with removal targets, a mask, and the ground truth.}
   \vspace{-1em}
   \label{fig:dataconstruct}
\end{figure*}


\section{SmartEraser}
The goal of object removal is to completely erase the target object specified by a masked region and seamlessly generate consistent content based on the surrounding context. Most existing methods follow a ``mask-and-inpaint" paradigm, where the masked area is excluded from the input, and the model attempts to inpaint the missing content based on the unmasked regions. While these methods can successfully remove objects in simple scenarios, they often regenerate new objects or produce blurs and artifacts in the masked region in more complex scenes. 

In contrast, we propose a novel paradigm: \textit{Masked-Region Guidance}. Instead of discarding masked regions, we retain it in the input, leveraging it to guide the removal process. This approach is based on the insight that, by knowing the exact removal targets, the model can precisely erase them while leaving the surrounding context maintained. In the following sections, we will first introduce the method of Masked-Region Guidance in Sec.~\ref{sec:mask_region_guide}. Then, we present Syn4Removal, a large-scale synthetic dataset designed for this paradigm in Sec.~\ref{sec:Syn4Removal_data_generation}. Finally, we describe \emph{SmartEraser}, a fine-tuned diffusion model based on the Mask-Region Guidance paradigm and Syn4Removal dataset in Sec.~\ref{sec:smarteraser_framework}.

\subsection{Masked-Region Guidance}
\label{sec:mask_region_guide}
Given an input image $\bm{x}$, and user-specified mask $\bm{m}$ containing the removal targets, the object removal task aims to produce the removal result $\bm{\hat{x}}$ while maintaining the realism of the original image. The mask $\bm{m}$ is a binary mask where 1 indicates the region containing the removal targets.
Existing models~\cite{lama,latentdm,clipaway,powerpaint} are mostly based on the ``mask-and-inpaint" paradigm, \textit{i.e.}, utilizing the unmasked region $\bm{x}\odot(\bm{1}-\bm{m})$ and mask $\bm{m}$ to predict the results without removal object in the masked region. These models are learned to maximize the following objective:
\vspace{-0.5em}
\begin{align}
    \mathcal{L}'=P(\bm{\hat{x}} | \bm{m},\bm{x}\odot(\bm{1}-\bm{m});\Theta '),
\end{align}
where the conditional probability $P$ is modeled using a neural
network with parameters $\Theta '$, $\odot$ specifies the element-wise multiplication.

In contrast, we introduce \textit{Masked-Region Guidance}, a fundamentally different approach. Instead of discarding the masked region, we retain it in the input, allowing it to guide the removal process. Specifically, we replace the condition input from $[\bm{m},\bm{x}\odot(\bm{1}-\bm{m})]$ to $[\bm{m},\bm{x}]$. The mask $m$ is still applied to instruct the region for removal targets, but $\bm{x}$ now includes the masked region $\bm{x} \odot \bm{m}$  besides $\bm{x}\odot(\bm{1}-\bm{m})$. By including the masked region, we provide the model with valuable guidance during the object removal task. \textit{First}, retaining the masked region allows the model to know exactly which objects need to be removed, reducing the risk of regenerating the object in the masked area. This helps the model focus on learning how to reconstruct the scene without the object. \textit{Second}, the masked region is typically larger than the removal targets. This allows the model to copy the surrounding context of target objects in the masked region to the final result, significantly reducing the difficulty of synthesizing a larger area. Moreover, if the masked region contains complex textures, these can be retained, leading to more realistic results. In our Masked-Region Guidance paradigm, object removal is learned to maximize the objective: 
\vspace{-0.5em}
\begin{align}
    \mathcal{L}=P(\bm{\hat{x}} | \bm{m},\bm{x};\Theta),
\end{align}
where the model is with parameters $\Theta$.

\subsection{Syn4Removal}
\label{sec:Syn4Removal_data_generation}


Now we have a new paradigm for training object removal models, but it cannot be directly applied using existing data construction methods. Typically, training data is created by masking parts of an image and having the model predict the masked content. However, this could lead the model to simply copy the masked content if applied to our paradigm. To overcome this, we require a dataset with triplets (input image, mask, and ground truth). However, existing real image datasets~\cite{rord, objectdrop} are limited in terms of scene variety, as they require photos taken before and after placing objects. Additionally, some synthetic datasets~\cite{defacto,instinpainting} rely on inpainting models to generate pseudo-removed results, which don't fully capture the real removal process. 

To address this, we propose a synthetic method for constructing object removal training data. Our approach involves pasting object instances from various images onto different backgrounds to form the input images. The mask of the pasted object serves as the input mask, and the original background image serves as the ground truth. Since the ground truth is the real image, it helps the model better capture the true distribution of the image, leading to more realistic object removal results. As a result, we constructed Syn4Removal, a large-scale high-quality data set that contains 1 million samples. The data construction pipeline is shown in Figure~\ref{fig:dataconstruct}. Next, we will introduce each component of the overall pipeline.



\noindent\textbf{Instance Crop \& Filter.} 
We directly use a public instance segmentation dataset~\cite{openimage} to get instances. We use the mask to get each instance from the image.
While cutting instances, we also compute the corresponding CLIP similarity score with their classes and normalized area ratio in the original images. Then we perform the filter process,
First, we exclude instances with areas larger than 95\% or smaller than 5\% to prevent potential low-quality segmentation results. 
Then, based on the previously calculated CLIP scores, we remove instances with low semantic relevance in each class. 
Given the variability in CLIP score distribution across different classes, we set individual CLIP score thresholds for each class according to previous work~\cite{xpaste}, filtering out instances with scores below these thresholds.

\begin{figure*}[t]
  \centering
   \includegraphics[width=1\linewidth]{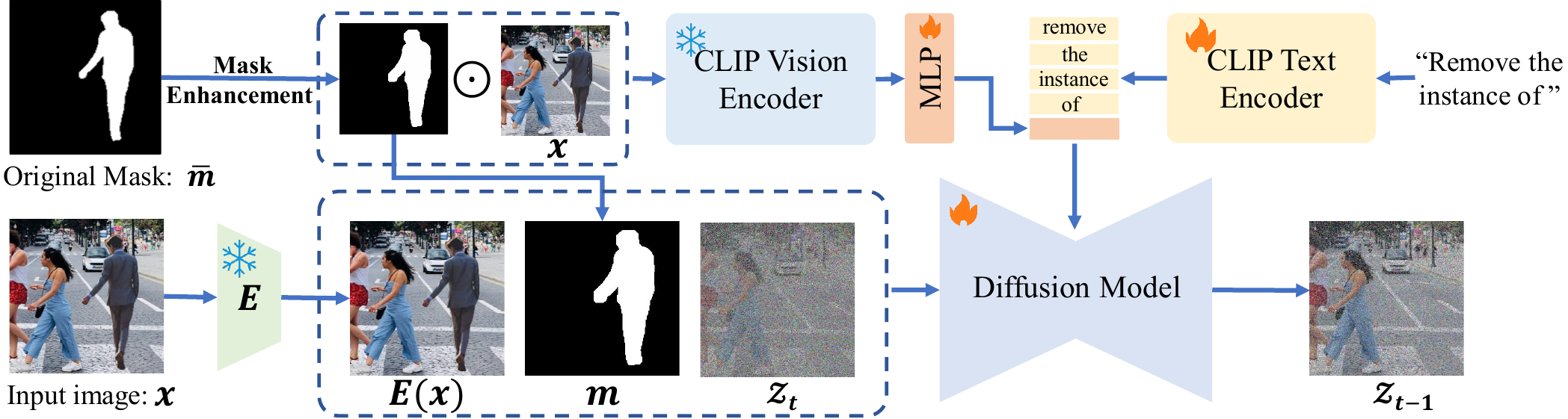}
   \vspace{-2em}
   \caption{Overall framework of SmartEraser. It fully utilizes the \textit{Masked-Region Guidance} paradigm and the Syn4Removal dataset.}
   \vspace{-1em}
   \label{fig:ModelFramework}
\end{figure*}

\noindent\textbf{Background Filter.}
For background images, we use public datasets COCONut~\cite{coconut} and SAM-1B~\cite{sam}, along with segmentation annotations. We filter out low-resolution images, those with extreme aspect ratios, or those with a large number of segmented instances. The selected backgrounds are then combined to form 1 million high-quality images (for more details, see Supplementary).

\noindent\textbf{Instance Size Sampling \& Resize.}
To match the size of the instance with that of the background, we first calculate the average $\mu_c$ and variance $\sigma^2_c$ of the normalized area ratio for each class $c$. When pasting an instance onto a background, we sample scale $s$ from a Gaussian distribution $\mathcal{N}(\mu_c, \sigma^2_c)$, and scale the instance to $sHW$, where $H$ and $W$ are the height and width of the background image, respectively.

\noindent\textbf{Feasible Area for Pasting.} To ensure that the pasted instance does not cover existing instances in the background image, we calculate IoU between the pasted instance region and each instance in the background image, the feasible region based on IoU results is formulated as:
\vspace{-0.5em}
\begin{align}
    \mathbf{R}_{1}=\{(x,y)\, |\, \text{IoU}(P_{x,y}, I_i) < r,\, I_i \in [1,I_{max}] \},
\end{align}
where $P_{x,y}$ is the instance region if centered on $(x,y)$, $r$ is a IoU threshold, $I_{max}$ is the number of existing instances.
Besides, to ensure the integrity of pasted instances, edge areas of background images are avoided, formulated as:
\vspace{-0.5em}
\begin{align}
        \mathbf{R}_{2}=\{(x,y)|x \in [e_x, W-e_x], y \in [e_y, H-e_y] \},
\end{align}
where $e_x$ and $e_y$ are half of the width and height of the bounding box of the instance mask, respectively. We take the intersection region of $\mathbf{R}_{1}$ and $\mathbf{R}_{2}$, the final set of feasible locations is as follows:
\vspace{-0.5em}
\begin{align}
        \mathbf{R}_{\text{f}} = \mathbf{R}_{1} \cap \mathbf{R}_{2}.
\end{align}
We then randomly select a location from $\mathbf{R}_{\text{f}}$ as the center point for pasting the instance. Based on the mask of the instance and location, we can get the instance mask centered on the location $\bm{\bar{m}}$. 

\noindent\textbf{Alpha Blending for Instance Pasting.}
To further ensure that the pasted instance blends harmoniously with the background, we apply alpha blending~\cite{alphamatting}. This method interpolates between the foreground instance and the background image, formulated as follows:
\vspace{-0.5em}
\begin{align}
\bm{x}=\boldsymbol \alpha \odot \bm{x_i} + (1-\boldsymbol{\alpha}) \odot \bm{x_b},
\end{align}
where $\bm{x}$ is final constructed image, $\bm{x_i}$ is the instance, $\bm{x_b}$ is the background image, and $\boldsymbol{\alpha}$ is alpha blending value. $\boldsymbol{\alpha}$ is calculated through an alpha blending algorithm~\cite{alphamatting} that processes the mask $\bm{\bar{m}}$  to a trimap and then takes the trimap as input, which consists of three regions, \textit{i.e.}, the foreground and background, and an excessively uncertain boundary region. So we have the triplet for object removal training: \{$\bm{x}$, $\bm{\bar m}$, $\bm{x_b}$\}, where $\bm{x}$ is input image with target object,   $\bm{\bar m}$ is original mask, and $\bm{x_b}$ as the ground truth. Using this data generation pipeline, we finally get about 1 million triplets that form the Syn4Removal dataset.


\subsection{Framework}
\label{sec:smarteraser_framework}
Based on the proposed Masked-Region Guidance paradigm and the Syn4Removal dataset, we design a framework based on the text-to-image stable diffusion model for object removal. In this framework, we carefully design a user-friendly mask enhancement strategy and introduce CLIP-based visual guidance to further improve the capability of the model for object removal, as shown in Figure~\ref{fig:ModelFramework}.

\begin{figure}[t]
  \centering
  \vspace{-0.5em}
   \includegraphics[width=1\linewidth]{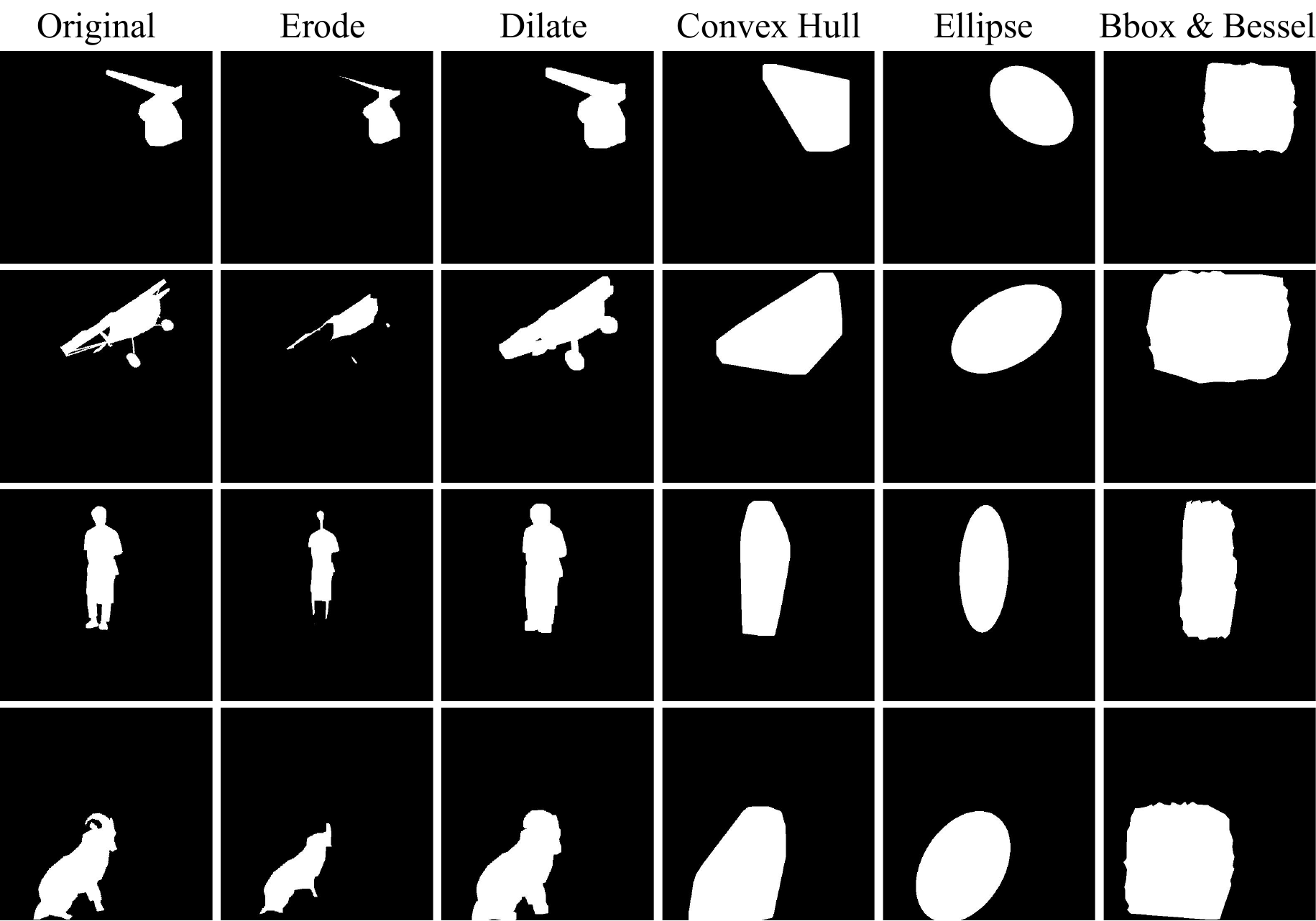}
   \vspace{-2em}
   \caption{Mask shapes from different mask enhancement methods.}
   \vspace{-1.5em}
   \label{fig:diff_mask_construct}
\end{figure}

\noindent\textbf{Mask Enhancement}. In real-world scenarios, users usually provide a loose or tight mask around the object to be removed. If the model were trained only with precise object masks, there would be a significant gap in mask shape and size between training and inference. To address this, during training, we apply various mask deformation methods to simulate the user input mask shapes, as shown in Figure~\ref{fig:diff_mask_construct}. These techniques help the model generalize to different mask forms. Specifically, we use six mask types to augment the object mask:
(1) \textit{Original mask}: this mask precisely outlines the target object. (2) \textit{Eroded mask}: created by applying morphological erosion to the original mask, simulating scenarios where the user’s mask may not fully cover the object. (3) \textit{Dilated mask}: created by applying morphological dilation to the original mask. (4) \textit{Convex hull mask}: constructed by calculating the convex hull that fully encompasses the original mask, then expanding it slightly. (5) \textit{Ellipse mask}: generated by finding the smallest enclosing ellipse around the original mask, followed by a slight expansion. (6) \textit{Bbox \& Bessel mask}: a bounding box is generated to completely cover the original mask first, then bessel curves are constructed along each edge of the box to introduce an irregular boundary. 
So the overall mask enhancement process is illustrated as follows:
\vspace{-0.3em}
\begin{equation}
    \bm{m}=\text{ME}_{i}(\bar{\bm{m}}),\quad{i\in\{1,...,6\}},
\end{equation}
where $\bm{m}$ is the enhanced mask used during training, $\text{ME}_{i}(\cdot)$ represents six different mask enhancement methods.

\begin{figure*}[t]
  \centering
   \includegraphics[width=1\linewidth]{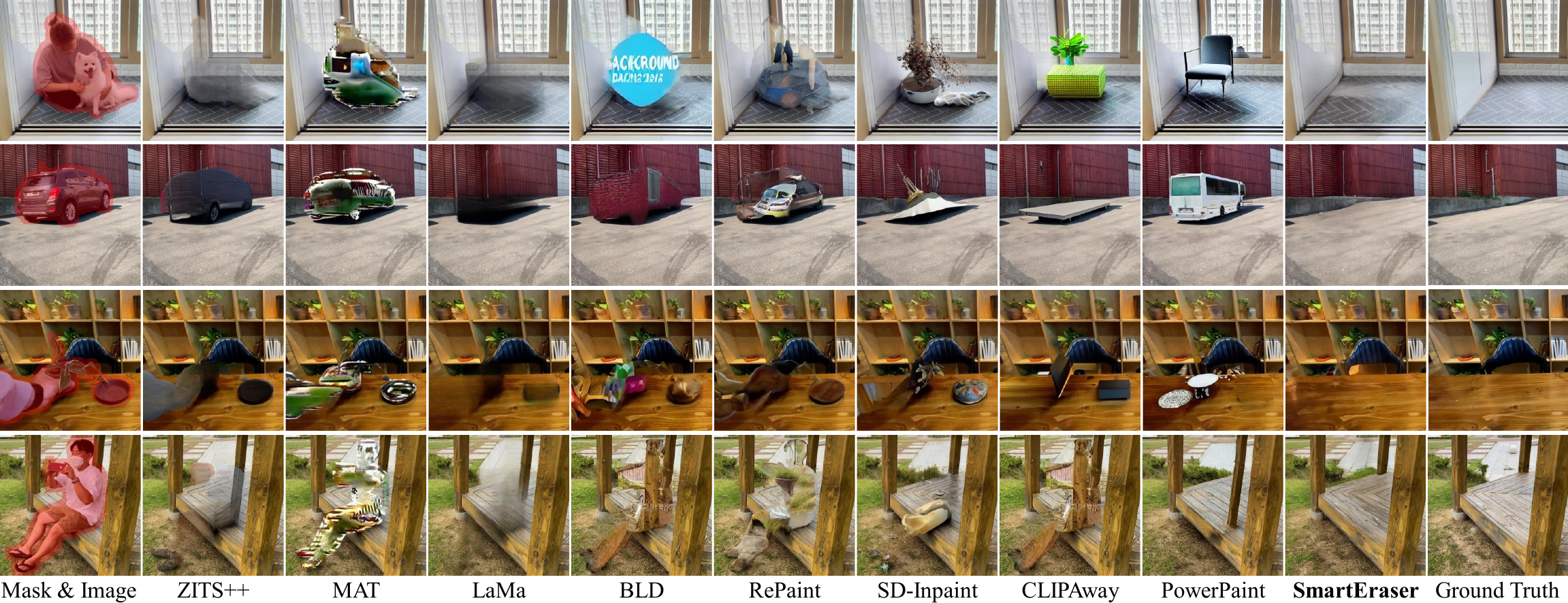}
   \vspace{-2em}
   \caption{Qualitative comparison of SmartEraser and state-of-the-art methods. The samples are sourced from RORD-Val, which includes the ground truth after removal.}
   \vspace{-1em}
   \label{fig:quantitative_experiment}
\end{figure*}

\noindent\textbf{CLIP-Based Visual Guidance}.
Our SmartEraser is designed based on text-to-image diffusion model, the text prompts play an important role in guiding the generation process. We aim to design a prompt that indicates the model what to remove during the generation process. Concretely, we first segment the region corresponding to the removal target in the image using its corresponding enhanced mask. This segmented region is then fed into the pre-trained CLIP~\cite{clip} vision encoder to extract its visual feature(\eg, class token). This feature is not precisely aligned with the text encoder output space. So we then apply a trainable MLP network to map the visual feature into the feature space of the text encoder. The mapped embedding is then appended to the tokens of the CLIP text embedding of the prompt ``\emph{Remove the instance of}''. 
During the training process, the vision encoder is frozen, and the MLP and text encoder are trainable. So the CLIP-based visual guidance is formulated as follows:
\vspace{-0.5em}
\begin{equation}
    \widehat{\bm{c}}=[\boldsymbol{\tau}_\theta(\bm{y}),\text{MLP}(\boldsymbol{\nu}_\theta(\bm{x}\odot \bm{m}))],
\end{equation}
where $\boldsymbol{\tau}_\theta(\cdot)$ is text encoder, $\boldsymbol{\nu}_\theta(\cdot)$ is vision encoder, and $\bm{y}$ represents input text prompt ``\emph{remove the instance of}''.

\noindent\textbf{Loss Function.} Suppose the $\bm{\textit{E}}(\cdot)$ is the VAE encoder, so the latent feature of input image $\bm{x}$ and its ground truth $\bm{x_b}$ is $\bm{\bar z} = \bm{\textit{E}}(x)$ and $\bm{z} = \bm{\textit{E}}(x_b)$, respectively, and $\bm{z}_t$ is the noisy feature of $\bm{z}$ with added noise $\epsilon$ at timestep $t$. So the overall training loss following the standard diffusion process, which is formulated as follows:
\vspace{-0.5em}
\begin{align}    
\mathcal{L}=\mathbb{E}_{\bm{x}, \bm{m}, \bm{x_b} ,t,\boldsymbol{\epsilon }}\|\boldsymbol{\epsilon}-\boldsymbol{\epsilon}_\theta(\bm{z}_t,\bm{\bar z}, \bm{m}, \widehat{\mathbf{c}},t)\|_2^2,
\end{align}
where $\boldsymbol{\epsilon}_\theta(\cdot)$ is the diffusion model.

\section{Experiments}

\begin{figure*}[t]
  \centering
   \includegraphics[width=1\linewidth]{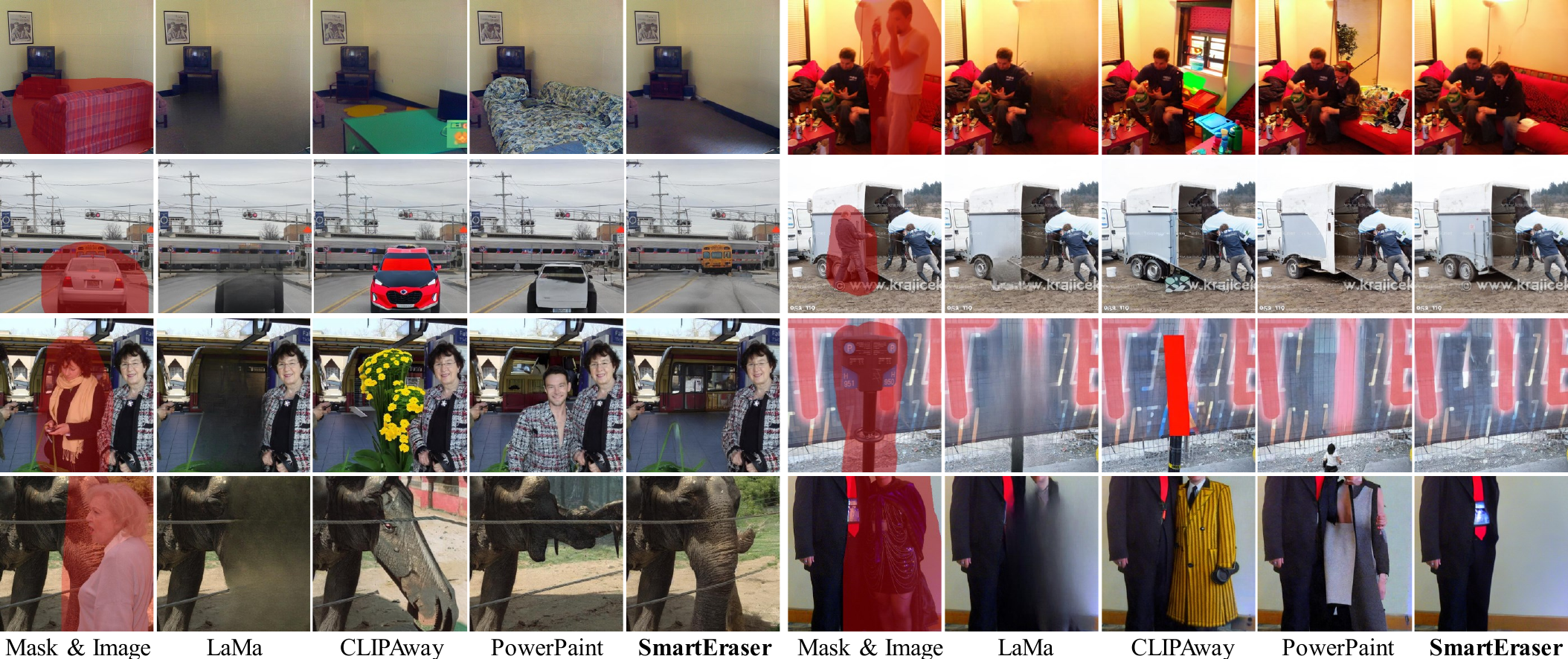}\
   \vspace{-2em}
   \caption{Qualitative comparison of SmartEraser and other methods in real-world user cases, in which users provide various mask shapes. The samples are sourced from real-world images and the validation set of MSCOCO~\cite{coco}.}
   \vspace{-1.5em}
   \label{fig:different_mask_experiment}
\end{figure*}

\subsection{Settings}
\textbf{Baselines.} We compare SmartEraser with methods in both image inpainting and object removal. For image inpainting, we evaluate against both GAN-based approaches (ZITS++~\cite{zits}, MAT~\cite{mat}, LaMa~\cite{lama}, BLD~\cite{bld}) and diffusion-based methods (RePaint~\cite{repaint}, and SD-Inpaint~\cite{latentdm}). Additionally, we compare it with specialized object removal frameworks, including CLIPAway~\cite{clipaway} and PowerPaint~\cite{powerpaint}, which represent the previous state-of-the-art performance on object removal benchmarks.


\noindent\textbf{Implementation details.} To enable a fair comparison, we follow previous experimental settings~\cite{clipaway,powerpaint} and finetune SmartEraser on the widely-used SD v1.5 model~\cite{latentdm}. The training process involves fine-tuning both the UNet and text encoder components within the diffusion model and MLP for mapping visual features to text features space. SmartEraser is trained on the proposed Syn4Removal dataset with a batch size of 32, using the AdamW~\cite{loshchilov2017decoupled} optimizer for 500$k$ iterations and a learning rate of 2e-5 across all trainable modules. 


\noindent\textbf{Evaluation benchmarks.} 
We evaluate the object removal capability of all baseline methods and our approach across three benchmarks in diverse scenarios. The benchmarks are constructed as follows:
\begin{itemize}    
    \item RORD-Val: Due to there are many triplets share the same background images as the ground truth in validation set of the RORD~\cite{rord}. To ensure unique scenes and high object quality, we filter this subset to obtain RORD-Val, a evaluation benchmark of real-photography image pairs.
    \item DEFACTO-Val: Derived from the splicing section of DEFACTO~\cite{defacto}, we exclude examples where intented targets for removal significantly overlap with instances in the background images.
    \item Syn4Removal-Val: Generated using the pipeline of Syn4Removal with MSCOCO~\cite{coco} validation set as background images and OpenImages-v7~\cite{openimage} validation set offering instances.
\end{itemize}
Each benchmark comprises approximately 5k samples. During inference, following~\cite{clipaway}, we apply slight mask dilation to mitigate downscaling artifacts in SD-Inpaint-based methods~\cite{latentdm, clipaway, powerpaint}.

\noindent\textbf{Evaluation metrics.} To quantitatively assess the performance of the object removal models, we consider three key aspects: (1) \textit{overall image quality}, which is evaluated by Fréchet Inception Distance~(FID)~\cite{fid} and CLIP Maximum Mean Discrepancy~(CMMD)~\cite{jayasumana2024rethinking}. (2) \textit{consistency between the predicted region and the background context}, evaluated by REMOVE~\cite{remove} metric. (3) \textit{consistency between the predicted region and corresponding region in the ground truth}, assessed by LPIPS~\cite{zhang2018unreasonable}, SSIM~\cite{ssim}, and PSNR~\cite{psnr}. 

\subsection{Comparasion with Previous Methods}

\textbf{Quantitative results.} 
We conduct extensive experiments to evaluate the performance of SmartEraser against baselines on the benchmarks.
The results are shown in Tables~\ref{tab:rord_result_sota}, \ref{tab:defacto_result_sota} and \ref{tab:paste_result_sota}. 
SmartEraser significantly outperforms all other methods across all metrics on the three benchmarks. Notably, on the more complex RORD-Val dataset, which has larger masks and more challenging removal scenes, SmartEraser surpasses the previous state-of-the-art by margins of 10.3 in FID and 1.89dB in PSNR. This indicates the superior performance of SmartEraser for object removal. Besides, SmartEraser also achieves state-of-the-art performance on the simpler DEFACTO-Val and Syn4Removal-Val datasets.


\begin{table}[t]
    \footnotesize
    \centering
    \resizebox{1.0\linewidth}{!}{
    \setlength\tabcolsep{0.4pt}
    \begin{tabular}{@{}lcccccc@{}}
        \toprule
        Method & FID~$\downarrow$ & CMMD~$\downarrow$ & ReMOVE~$\uparrow$ & LPIPS~$\downarrow$ & SSIM~$\uparrow$ & PSNR~$\uparrow$ \\
        \midrule
        ZITS++~\cite{zits}   & 53.440 & 0.838 & 0.820 & 0.388 & 0.522 & 15.442 \\
        MAT~\cite{mat}  & 80.953 & 1.494 & 0.628 & 0.433 & 0.495 & 13.322  \\
        LaMa~\cite{lama} & 24.237 & 0.216 & 0.916 & 0.348 & 0.557 & 16.383 \\
        BLD~\cite{bld}    & 73.086 & 1.760 & 0.834 & 0.471 & 0.467 & 15.775 \\
        RePaint~\cite{repaint} & 71.542 & 1.639 & 0.812 & 0.381 & 0.510 & 15.861 \\
        SD-Inpaint~\cite{latentdm}    & 69.502 & 0.324 & 0.857 & 0.369 & 0.537 & 16.111 \\
        CLIPAway~\cite{clipaway}    & 25.458 & 0.123 & 0.915 & 0.333 & 0.577 & 17.434 \\
        PowerPaint~\cite{powerpaint}    & 24.058 & 0.294 & 0.926 & 0.308 & 0.602 & 18.100 \\
        \textbf{SmartEraser} & \textbf{16.030} & \textbf{0.092} & \textbf{0.937} & \textbf{0.276} & \textbf{0.612} & \textbf{19.994}  \\
        \bottomrule
    \end{tabular}
    }
    \vspace{-1em}
    \caption{Quantitative comparison of SmartEraser and previous methods on RORD-Val.}
    \vspace{-2em}
    \label{tab:rord_result_sota}
\end{table}

\begin{table}[t]
    \footnotesize
    \centering
    \resizebox{1.0\linewidth}{!}{
    \setlength\tabcolsep{0.4pt}
    \begin{tabular}{@{}lcccccc@{}}
        \toprule
        Method & FID~$\downarrow$ & CMMD~$\downarrow$ & ReMOVE~$\uparrow$ & LPIPS~$\downarrow$ & SSIM~$\uparrow$ & PSNR~$\uparrow$ \\
        \midrule
        ZITS++~\cite{zits}   & 8.153 & 0.229 & 0.899 & 0.350 & 0.634 & 19.453 \\
        MAT~\cite{mat}    & 15.257 & 0.321 & 0.742 & 0.407 & 0.575 & 14.541 \\
        LaMa~\cite{lama}       & 6.601 & 0.201 & 0.926 & 0.351 & 0.683 & 21.735 \\
        BLD~\cite{bld}    & 24.407 & 0.746 & 0.913 & 0.418 & 0.632 & 19.346 \\
        RePaint~\cite{repaint}    & 12.314 & 0.822 & 0.924 & 0.325 & 0.644 & 20.607 \\
        SD-Inpaint~\cite{latentdm}    & 11.119 & 0.196 & 0.912 & 0.320 & 0.656 & 21.691 \\
        CLIPAway~\cite{clipaway}    & 8.320 & 0.193 & 0.925 & 0.323 & 0.666 & 21.953 \\
        PowerPaint~\cite{powerpaint}    & 6.582 & 0.275 & 0.934 & 0.278 & 0.715 & 23.870 \\
        \textbf{SmartEraser} & \textbf{3.405} & \textbf{0.106} & \textbf{0.939} & \textbf{0.257} & \textbf{0.734} & \textbf{25.363}  \\
        \bottomrule
    \end{tabular}
    }
    \vspace{-1em}
    \caption{Quantitative comparison of SmartEraser and previous methods on DEFACTO-Val.}
    \vspace{-2em}
    \label{tab:defacto_result_sota}
\end{table}

\begin{table}[t]
    \footnotesize
    \centering
    \resizebox{1.0\linewidth}{!}{
    \setlength\tabcolsep{0.4pt}
    \begin{tabular}{@{}lcccccc@{}}
        \toprule
        Method & FID~$\downarrow$ & CMMD~$\downarrow$ & ReMOVE~$\uparrow$ & LPIPS~$\downarrow$ & SSIM~$\uparrow$ & PSNR~$\uparrow$ \\
        \midrule
        ZITS++~\cite{zits}   & 5.932 & 0.193 & 0.810 & 0.358 & 0.581 & 17.315 \\
        MAT~\cite{mat}    & 19.263 & 0.300 & 0.700 & 0.374 & 0.570 & 13.357 \\
        LaMa~\cite{lama}       & 5.246 & 0.097 & 0.932 & 0.325 & 0.606 & 19.319 \\
        BLD~\cite{bld}    & 20.286 & 0.705 & 0.896 & 0.464 & 0.525 & 16.539 \\
        RePaint~\cite{repaint}    & 12.677 & 0.696 & 0.908 & 0.378 & 0.553 & 17.689 \\
        SD-Inpaint~\cite{latentdm}    & 11.528 & 0.107 & 0.897 & 0.348 & 0.572 & 18.105 \\
        CLIPAway~\cite{clipaway}    & 5.530 & 0.093 & 0.930 & 0.344 & 0.593 & 18.561 \\
        PowerPaint~\cite{powerpaint}    & 5.025 & 0.200 & 0.931 & 0.291 & 0.663 & 20.594 \\
        \textbf{SmartEraser} & \textbf{4.386} & \textbf{0.053} & \textbf{0.939} & \textbf{0.269} & \textbf{0.672} & \textbf{22.029} \\
        \bottomrule
    \end{tabular}
    }
    \vspace{-1em}
    \caption{Quantitative comparison of SmartEraser and previous methods on Syn4Removal-Val.}
    \vspace{-1em}
    \label{tab:paste_result_sota}
\end{table}


\noindent\textbf{Qualitative results.} We present visualization comparison results of object removal using images from real scenes, shown in Figure~\ref{fig:quantitative_experiment}. 
Baseline methods have limitations and may introduce unintended objects, create unrealistic structures, or produce artifacts within masked regions. In contrast, SmartEraser succeeds in effectively removing target objects while generating high-quality synthesized results, and excels in  maintaining the contextual integrity and visual coherence of the scene.



\begin{table}[t]
    \footnotesize
    \centering
    \resizebox{1.0\linewidth}{!}{
    \setlength\tabcolsep{1pt}
    \begin{tabular}{@{}lccc@{}}
        \toprule
        Method & Image Quality~$\uparrow$ & FB-Consistency~$\uparrow$ & Removal Accuracy~$\uparrow$ \\
        \midrule
        LaMa~\cite{lama}   & 0.10 & 0.11 & 0.21 \\
        CLIPAway~\cite{clipaway}  & 0.18 & 0.12 & 0.11 \\
        PowerPaint~\cite{powerpaint} & 0.16 & 0.12 & 0.10 \\
        \textbf{SmartEraser}    & \textbf{0.56} & \textbf{0.65} & \textbf{0.58} \\
        \bottomrule
    \end{tabular}
    }
    \vspace{-1em}
    \caption{Comparison of user study results in terms of image quality, foreground-background consistency, and removal accuracy.}
    \vspace{-2em}
    \label{tab:user_study}
\end{table}

\noindent\textbf{User study.} To assess the visual quality subjectively, we conduct a user study with 30 participants. The study includes 20 image sets, each containing an original image, its mask, and removal results from four different methods, presented in random order. Participants are asked to select their preferred result based on three criteria: overall image quality, foreground-background consistency, and accuracy of object removal. Voting percentages in Table~\ref{tab:user_study} show that users consistently favor the results from SmartEraser.


\noindent\textbf{Real-world user cases.}
To further evaluate the object removal capability of SmartEraser in real-world user scenarios, we conduct a comparison with other methods (LaMa~\cite{lama}, CLIPAway~\cite{clipaway}, PowerPaint~\cite{powerpaint}), using masks provided by users. Figure~\ref{fig:different_mask_experiment} shows the comparison results. We can observe that SmartEraser exhibits a remarkable ability to accurately identify and remove target objects while preserving surrounding contextual details, particularly in scenarios involving imperfect or approximate masks. In the sample in the bottom row on the left, SmartEraser accurately identifies and removes the \emph{person} indicated by the mask, while maintaining the integrity of the \emph{elephant} in the background. In contrast, alternative methods tend to repaint the entire masked area, which leads to outputs that lack contextual coherence and visual realism due to the loss of details of the \emph{elephant}. The upper second sample on the right provides a challenging scenario, the user-provided mask aims to remove the \emph{man} whose feet are partially obscured by \emph{floating text}. The previous methods fail to maintain the integrity of the \emph{floating text}. However, SmartEraser  accurately removes the \emph{man} while preserving the \emph{floating text}.
 This highlights SmartEraser's ability to effectively differentiate between removal targets and background content, even in cases of partial occlusion.
 



\subsection{Ablation studies}
\begin{figure}[t]
  \centering
   \includegraphics[width=1\linewidth]{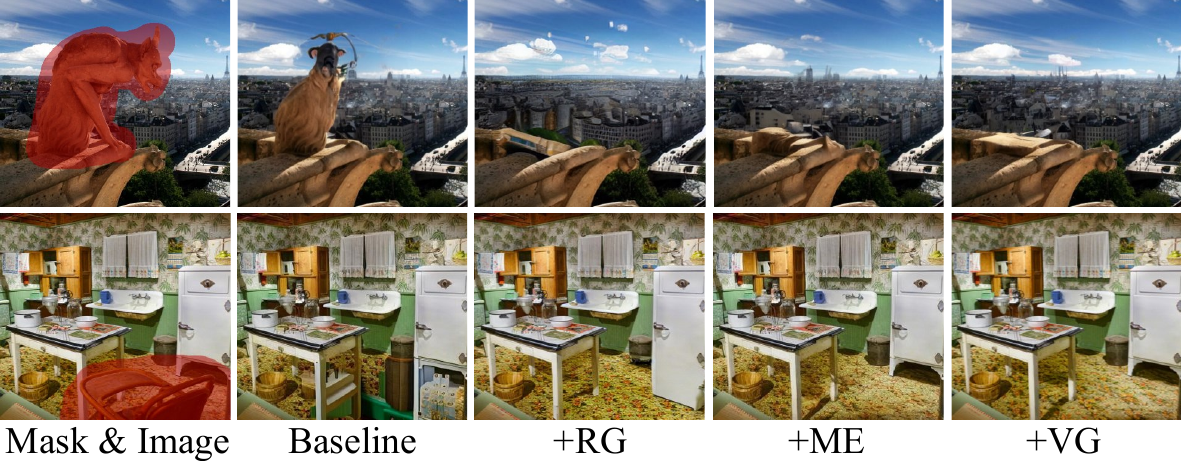}
   \vspace{-1.5em}
   \caption{Qualitative ablation comparison of our method. From left to right, we progressively add each proposed component.}
   \vspace{-0.5em}
   \label{fig:ablation_study}
\end{figure}

To evaluate the effectiveness of the key components in our framework, we perform ablation studies and report both quantitative and qualitative results. We progressively modify the baseline (fine-tuning SD v1.5 with the ``mask-and-inpaint'' paradigm) by adding our proposed techniques: RG for masked-region guidance, ME for mask enhancement, and VG for CLIP-based visual guidance. 
The results are displayed in Figure~\ref{fig:ablation_study} and Table~\ref{tab:effect_of_each_component_rord}. 

\noindent\textbf{Effect of the Masked-Region Guidance paradigm}: We observe that the baseline model often regenerates objects instead of properly removing them, as it is trained in a traditional ``inpainting" manner. Incorporating masked-region guidance significantly improves the removal performance by reducing the risk of object regeneration. However, due to the mismatch between the mask shape used during training and the user-defined mask during testing, the model does not always perfectly remove the target objects and preserve the surrounding context.

\noindent\textbf{Contribution of Mask Enhancement}: By simulating real-world user mask shapes over object removal during training, we find that it helps reduce the gap between model training and testing using user input mask, significantly keeping the surrounding regions and further improving removal results.

\noindent\textbf{Value of CLIP-based Visual Guidance}: CLIP visual guidance aids the model in better understanding the removal targets, yielding the best performance on RORD-val across all metrics. This indicates that the visual guidance helps the model perform object removal tasks more effectively.

\begin{table}[t]
    \footnotesize
    \centering
    \resizebox{1.0\linewidth}{!}{
    \setlength\tabcolsep{1.5pt}
    \begin{tabular}{@{}lcccccc@{}}
        \toprule
        Method & FID~$\downarrow$ & CMMD~$\downarrow$ & ReMOVE~$\uparrow$ & LPIPS~$\downarrow$ & SSIM~$\uparrow$ & PSNR~$\uparrow$ \\
        \midrule
        Baseline & 33.131 & 0.245 & 0.908 & 0.355 & 0.544 & 16.535 \\
        +RG & 24.526 & 0.121 & 0.923 & 0.339 & 0.562 & 17.176 \\
        \quad+ME & 19.421 & 0.108 & 0.931 & 0.301 & 0.591 & 18.917 \\
        \quad\quad+VG & 16.030 & 0.092 & 0.937 & 0.276 & 0.616 & 19.994 \\
        \bottomrule
    \end{tabular}
    }
    \vspace{-1em}
    \caption{Quantitative ablated comparison on RORD-Val.}
    \vspace{-2em}
    \label{tab:effect_of_each_component_rord}
\end{table}

\section{Conclusion}
In conclusion, we present \textit{SmartEraser}, a state-of-the-art object removal model built on the novel \textit{Masked-Region Guidance} paradigm. SmartEraser outperforms existing methods by smartly identifying the target object to remove while effectively preserving the surrounding context. To facilitate research on this paradigm, we propose Syn4Removal, a large-scale, high-quality dataset containing over a million image triplets, specifically designed for object removal tasks. Through extensive experiments, we demonstrate that SmartEraser achieves superior performance in both quality and robustness compared to previous object removal methods.




\appendix
\section{More Dataset Details}
In this section, we provide more implementation details about the proposed Syn4Removal dataset.

\noindent\textbf{Instance Filter.}
Following the previous work~\cite{xpaste}, considering the variability of CLIP score distribution across different instance classes, we compute thresholds $thres_{c}$ for each class $c$, which is formulated as:
\begin{align}
    thres_{c}=min(b, max(S_{c})-d), 
\end{align}
where $S_{c}$ is the CLIP score sets of all instances in the class $c$, $b$ and $d$ are predefined thresholds, set to 0.2 and 0.02, respectively. After that, we exclude instances whose CLIP scores fall below the threshold of their corresponding class.

\noindent\textbf{Background Filter.}
To obtain suitable background images from public datasets COCONut~\cite{coconut} and SAM-1B~\cite{sam} for instance pasting, we apply several filtering criteria. 
1) Background images are excluded if their width or height resolution is below 512 pixels, to prevent low-resolution images from being used as ground truths during the training process. 
2) Images with an aspect ratio larger than 2 are also discarded, to reduce the risk of image distortion during resizing and cropping to 512$\times$512 resolutions before inputting to SD v1.5. 
3) If the total area covered by instances in the background image exceeds 85\%, the image is also excluded, as it becomes challenging to compute a suitable region for pasting instances. 
After filtering, we obtain approximately 750$k$ background images from SAM-1B~\cite{sam}, and 282$k$ background images from COCONut~\cite{coconut}, resulting in about 1M training data.

\section{More Framework Details}
In this section, we provide more implementation details about the architecture of the MLP module in the proposed CLIP-based visual guidance.

\begin{figure}[h]
  \centering
   \includegraphics[width=1\linewidth]{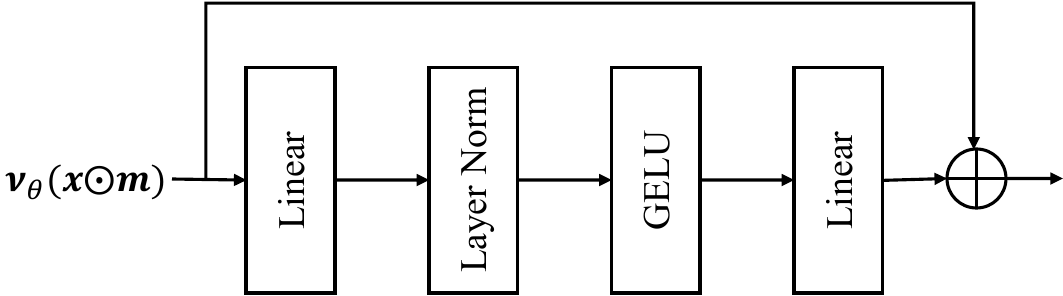}
   \vspace{-1.5em}
   \caption{The detailed architecture of the MLP module.}
   \vspace{-1.5em}
   \label{fig:mlp_detail}
\end{figure}

As shown in Figure~\ref{fig:mlp_detail}, the MLP module is designed with a simple but effective architecture. It is employed to map the visual feature extracted by the CLIP vision encoder into the feature space of the text encoder. It consists of two linear layers, a layer normalization layer, and a GELU~\cite{hendrycks2016gaussian} activation function. In addition, a residual connection is incorporated between the input and output to preserve the original feature generated by the CLIP visual encoder. The dimensions of the input, hidden state, and output are equal to 768.

\section{Samples from Syn4Removal}
\begin{figure*}[t]
  \centering
   \includegraphics[width=0.95\linewidth]{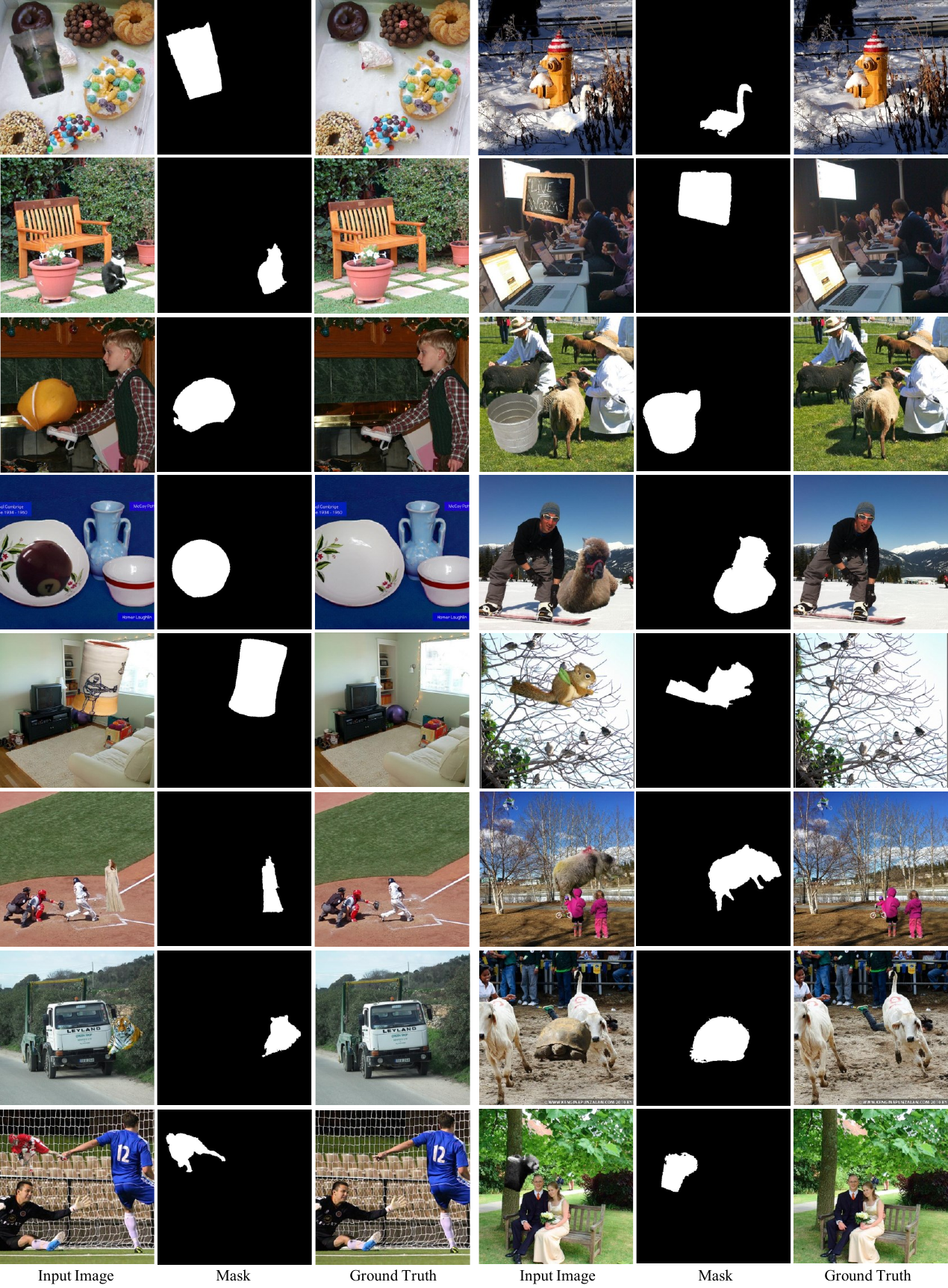}
   \vspace{-1em}
   \caption{Samples from the Syn4Removal dataset, including input images, masks, and ground-truth images.}
   \label{fig:data_sample}
\end{figure*}

In Figure~\ref{fig:data_sample}, we present examples from Syn4Removal, including input images, masks, and ground-truth images, with the carefully designed pipeline for creating triplets.
To prevent excessive overlap with existing instances in the background, we compute the Intersection over Union (IoU) between the pasted objects and each instance in the background, carefully determining the location of the pasted objects. Additionally, the pasted instances are seamlessly integrated into the background images by using alpha blending, ensuring visual harmony.
Our final triplet data is high-quality and more suitable for the object removal task.

\section{Additional Experiments}
In this section, we provide more comprehensive experiments with quantitative and qualitative results reported to demonstrate the effectiveness of our method.

\begin{figure}[t]
  \centering
   \includegraphics[width=1\linewidth]{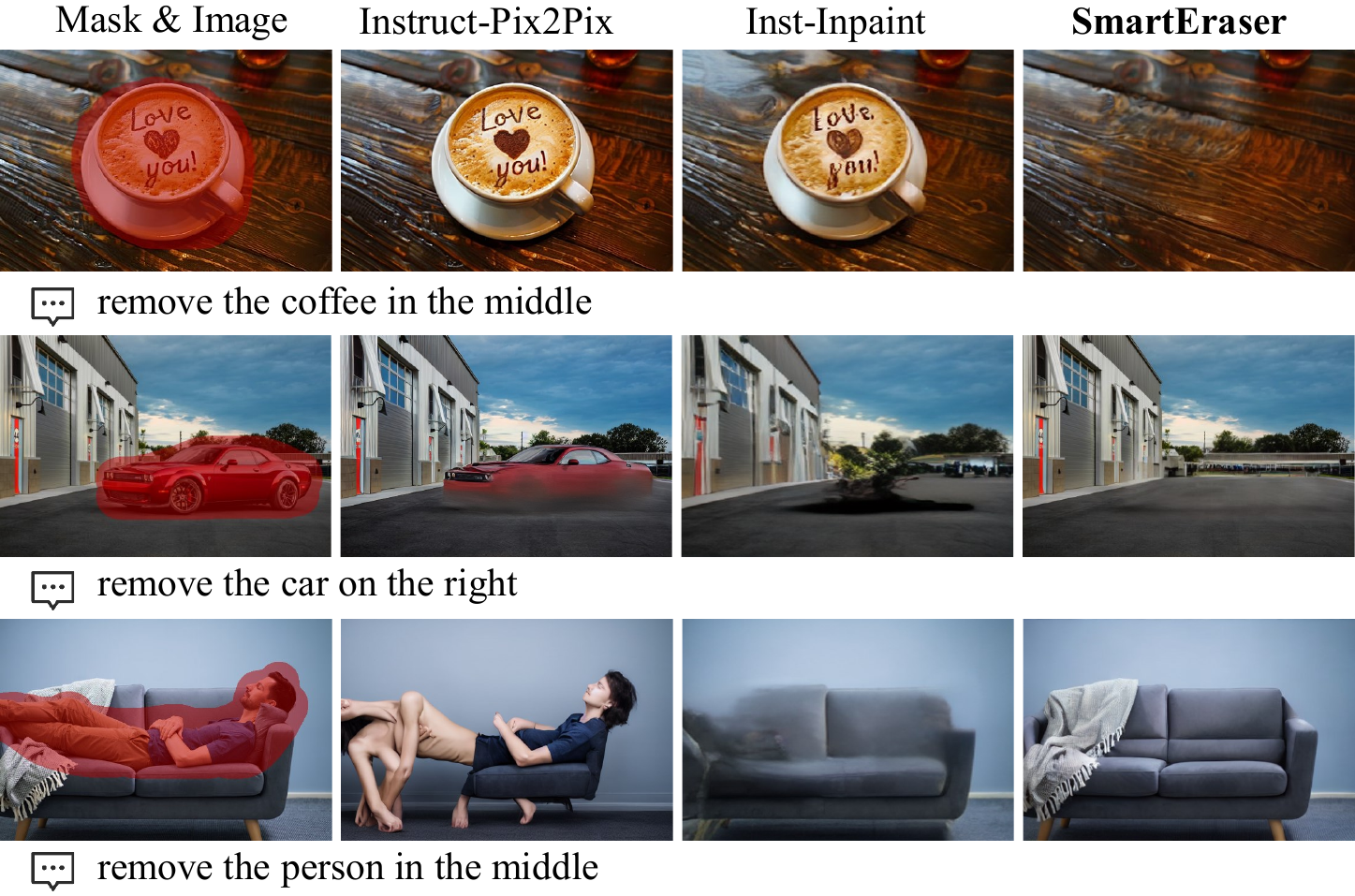}
   \vspace{-2em}
   \caption{Qualitative comparison of SmartEraser and existing instruction-based image editing method, which only rely on user input prompts for object removal.}
   \vspace{-0.5em}
   \label{fig:comp_instruction}
\end{figure}

\noindent\textbf{Comparisons with Instruction-Based Methods.}
In Figure \ref{fig:comp_instruction}, we provide qualitative comparisons between instruction-based methods~\cite{instructpix2pix,instinpainting} and our SmartEraser. Instruct-Pix2Pix~\cite{instructpix2pix} struggles in removing objects, often failing to remove the target objects, instead, introducing unrealistic edits to the objects. Another work Inst-Inpaint~\cite{instinpainting} demonstrates a basic capability for object removal, but it faces challenges in more complex cases, resulting in incomplete object removal and poor coherence between the removed area and the background, as observed in the first and second samples. Additionally, Inst-Inpaint has difficulty identifying the target object only based on user instructions. For instance, while the person is removed in the third sample, the stall on the sofa is also removed, which is contrary to the intention of the user.
Furthermore, since it is trained on synthetic ground truths, Inst-Inpaint tends to produce blurred generation. 
In contrast, SmartEraser effectively removes target objects accurately and preserves high background consistency and overall image quality.

\noindent\textbf{Ablation Studies on Other Benchmarks.}
We further supply the quantitative experiments of ablation studies on DEFACTO-Val and Syn4Removal-Val. As shown in Tables~\ref{tab:effect_of_each_component_defacto} and~\ref{tab:effect_of_each_component_syn}, the overall performance progressively improves as key techniques are incrementally integrated into the baseline. In the Tables, RG represents the masked-region guidance, ME denotes mask enhancement, and VG indicates CLIP-based visual guidance. These quantitative results demonstrate the effectiveness of each key component in our proposed framework.

\begin{table}[t]
    \footnotesize
    \centering
    \resizebox{1.0\linewidth}{!}{
    \setlength\tabcolsep{1.5pt}
    \begin{tabular}{@{}lcccccc@{}}
        \toprule
        Method & FID~$\downarrow$ & CMMD~$\downarrow$ & ReMOVE~$\uparrow$ & LPIPS~$\downarrow$ & SSIM~$\uparrow$ & PSNR~$\uparrow$ \\
        \midrule
        Baseline & 10.125 & 0.189 & 0.916 & 0.315 & 0.674 & 22.135 \\
        +RG & 7.142 & 0.132 & 0.932 & 0.279 & 0.712 & 24.176 \\
        \quad+ME & 4.701 & 0.125 & 0.938 & 0.271 & 0.721 & 24.917 \\
        \quad\quad+VG & 3.405 & 0.106 & 0.939 & 0.257 & 0.734 & 25.363 \\
        \bottomrule
    \end{tabular}
    }
    \vspace{-1em}
    \caption{Quantitative ablated comparison on DEFACTO-Val.}
    \vspace{-1em}
    \label{tab:effect_of_each_component_defacto}
\end{table}

\begin{table}[t]
    \footnotesize
    \centering
    \resizebox{1.0\linewidth}{!}{
    \setlength\tabcolsep{1.5pt}
    \begin{tabular}{@{}lcccccc@{}}
        \toprule
        Method & FID~$\downarrow$ & CMMD~$\downarrow$ & ReMOVE~$\uparrow$ & LPIPS~$\downarrow$ & SSIM~$\uparrow$ & PSNR~$\uparrow$ \\
        \midrule
        Baseline & 8.755 & 0.104 & 0.908 & 0.335 & 0.584 & 18.735 \\
        +RG & 5.419 & 0.081 & 0.927 & 0.299 & 0.632 & 20.176 \\
        \quad+ME & 4.748 & 0.067 & 0.934 & 0.281 & 0.661 & 21.317 \\
        \quad\quad+VG & 4.386 & 0.053 & 0.939 & 0.269 & 0.672 & 22.029 \\
        \bottomrule
    \end{tabular}
    }
    \vspace{-1em}
    \caption{Quantitative ablated comparison on Syn4Removal-Val.}
    \vspace{-1em}
    \label{tab:effect_of_each_component_syn}
\end{table}

\begin{figure}[t]
  \centering
   \includegraphics[width=1\linewidth]{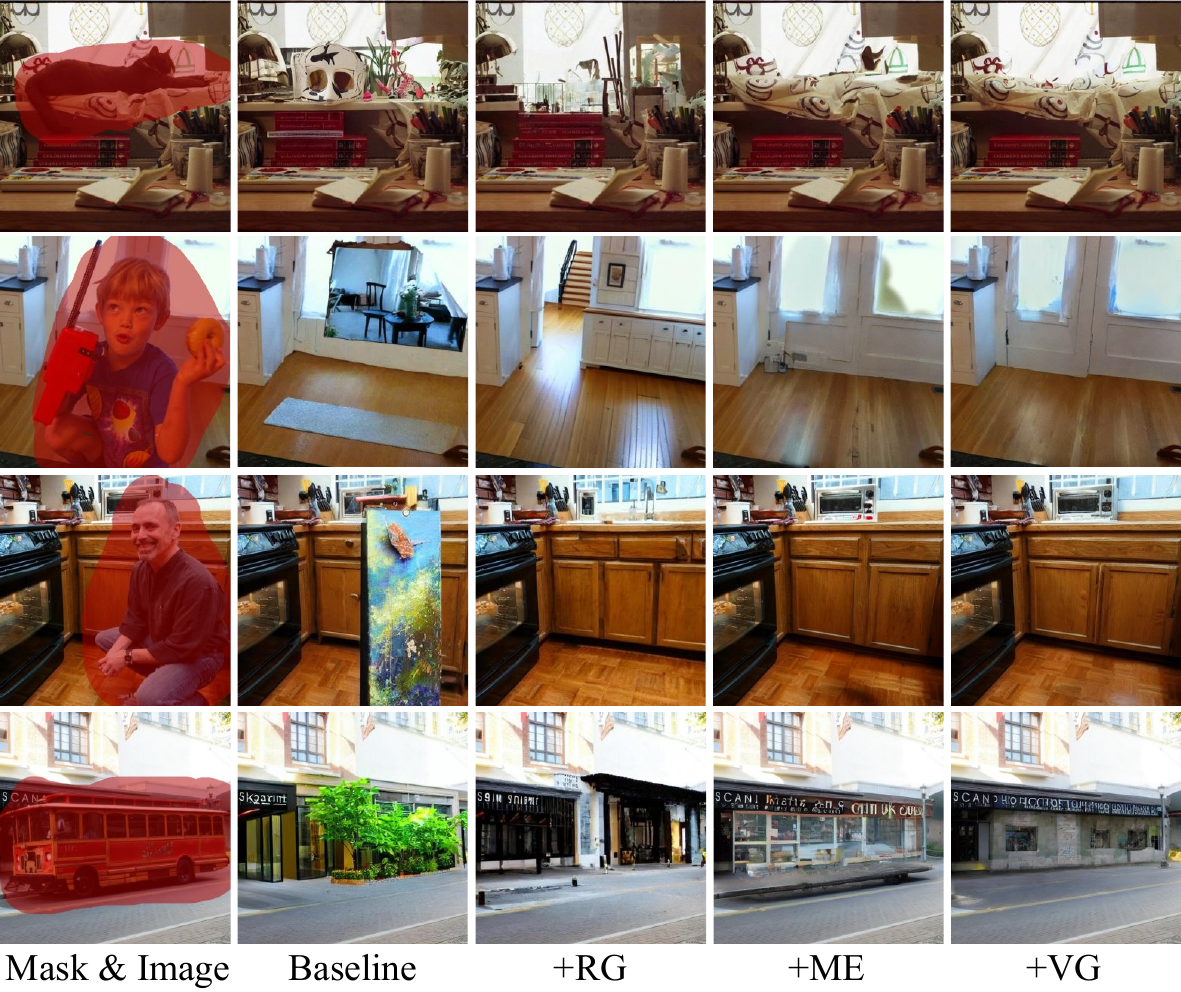}
   \vspace{-2em}
   \caption{Qualitative ablation comparison of our method. From left to right, we progressively add each proposed component.}
   \vspace{-0.5em}
   \label{fig:sup_ablation_study}
\end{figure}

\noindent\textbf{More Qualitative Results of Ablation Studies.}
As shown in Figure~\ref{fig:sup_ablation_study}, we provide additional qualitative results of ablation studies, with images sourced from the validation set of MSCOCO~\cite{coco}. These examples reveal that the baseline model, trained using the ``mask-and-inpaint" paradigm on the Syn4Removal dataset, frequently regenerates unintended objects within the masked regions. After adding the masked-region guidance, the risk of regeneration is significantly reduced. However, we observe that the background context within the masks around the target objects has also changed obviously. To address this, we introduce the mask enhancement to simulate the user-provided masks to reduce the gap between the training and inference, which effectively helps the model to preserve the surrounding context and further improve the removal results. Finally, integrating CLIP-based visual guidance (VG) provides explicit semantic guidance, enabling the model to achieve better object removal performance with more coherence and fidelity.

\begin{figure}[t]
  \centering
   \includegraphics[width=1\linewidth]{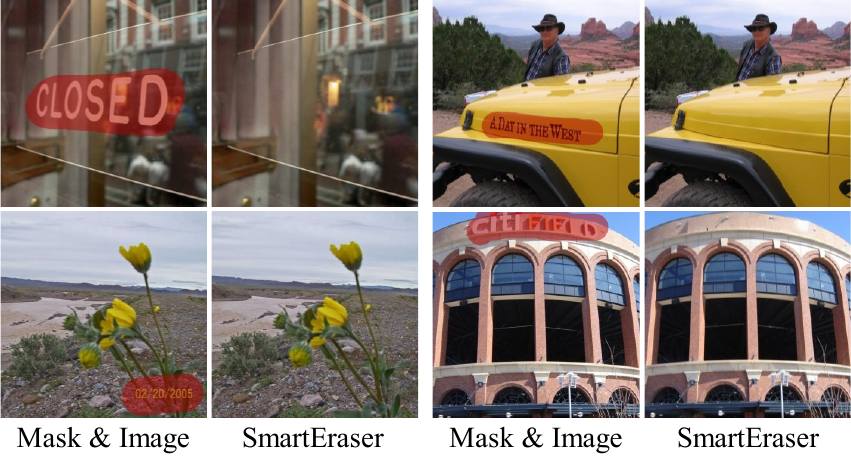}
   \vspace{-2em}
   \caption{Qualitative results of our method for erasing scene text.}
   \vspace{-1.5em}
   \label{fig:text_erase}
\end{figure}

\begin{figure*}[t]
  \centering
   \includegraphics[width=1\linewidth]{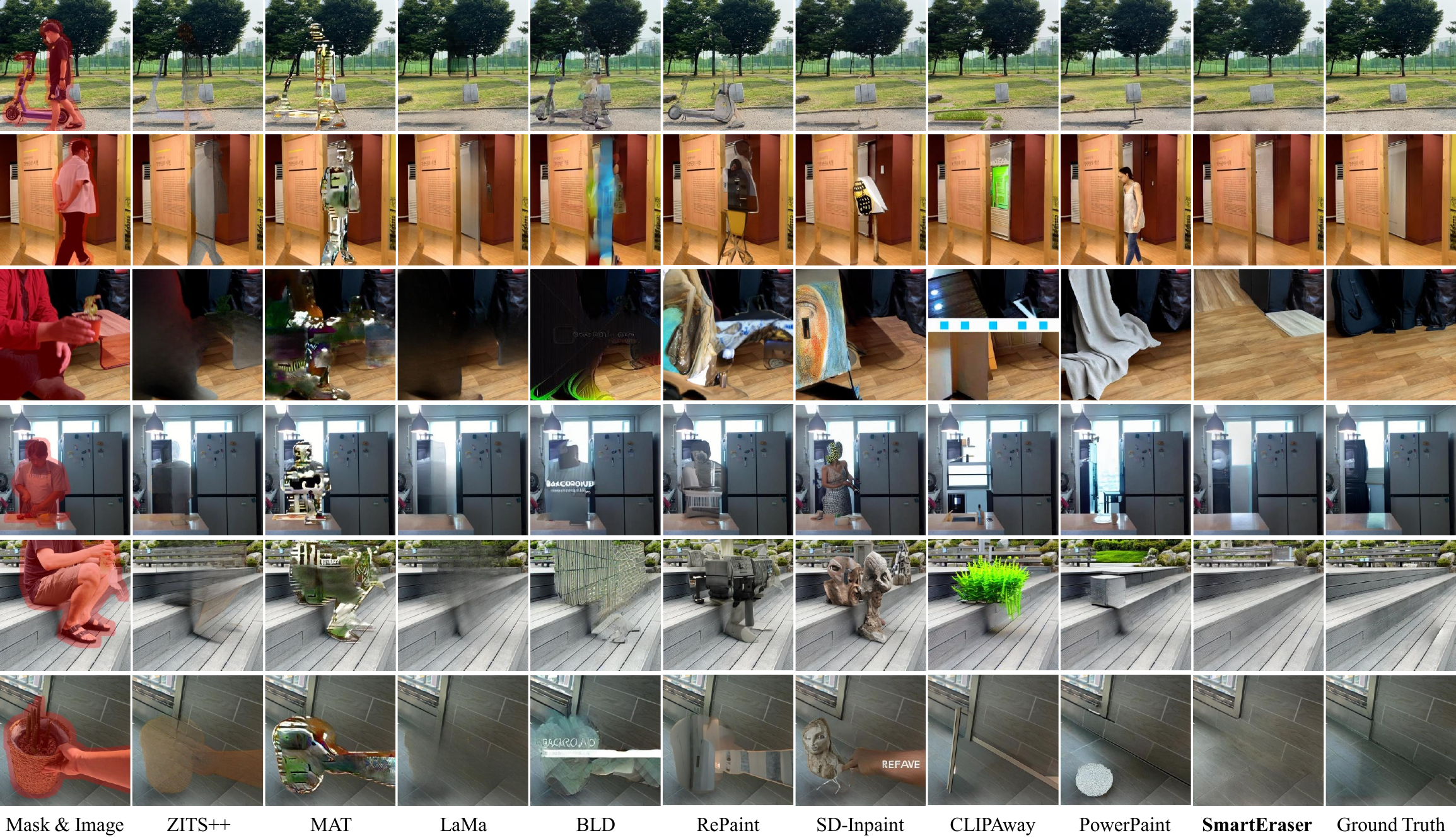}
   \vspace{-1em}
   \caption{Qualitative comparison of previous methods and SmartEraser. The samples are sourced from RORD-Val.}
   \vspace{-1em}
   \label{fig:more_quantitative_experiment_rord}
\end{figure*}

\begin{figure*}[t]
  \centering
   \includegraphics[width=1\linewidth]{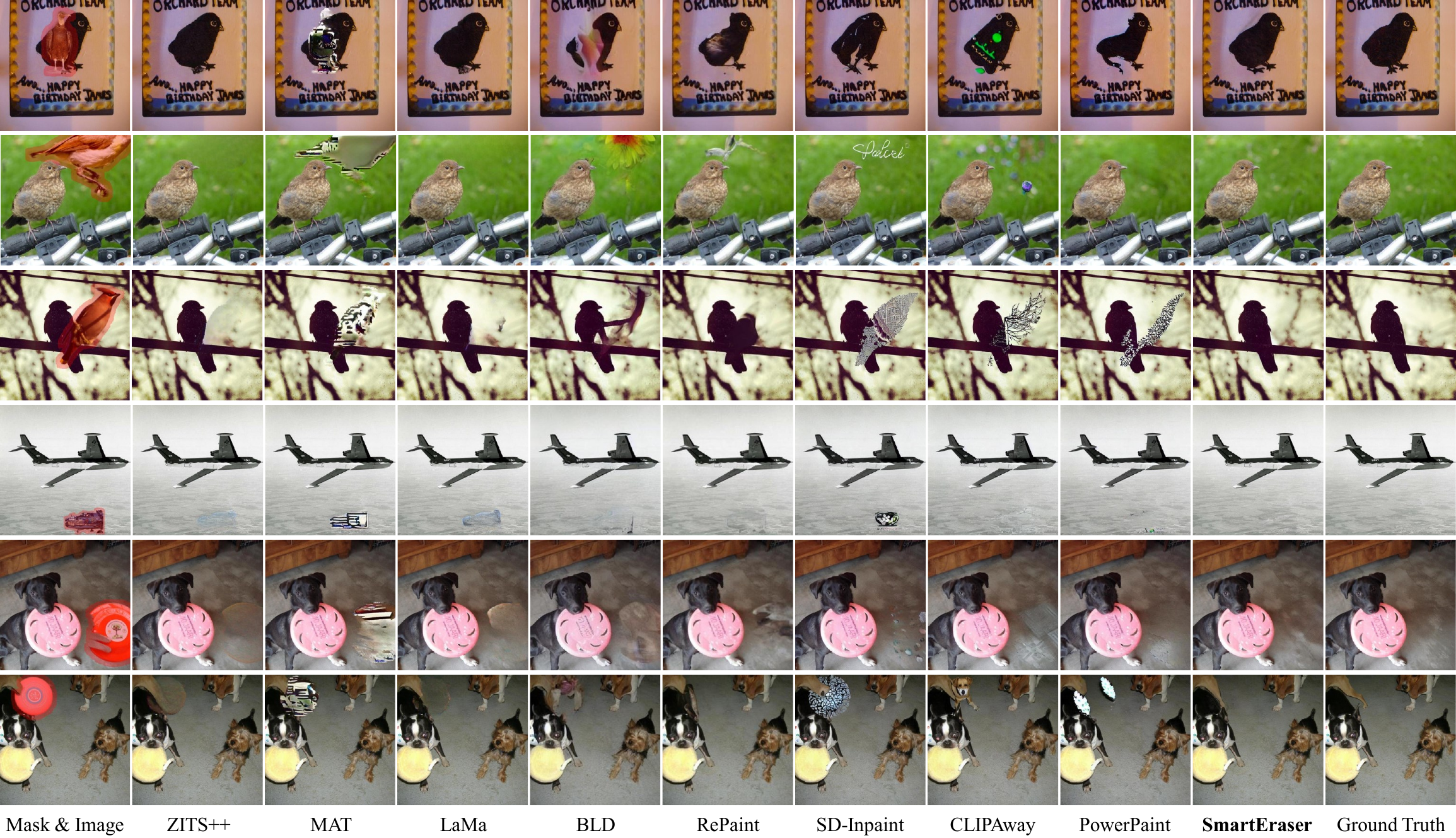}
   \vspace{-1em}
   \caption{Qualitative comparison of previous methods and SmartEraser. The samples are sourced from the splicing section in DEFACTO dataset.}
   \label{fig:more_quantitative_experiment_defacto}
   \vspace{-1em}
\end{figure*}

\noindent\textbf{More Qualitative Results with Previous Methods.}
To better express the excellent performance of SmartEraser in object removal, we present more qualitative comparisons with previous methods in Figures~\ref{fig:more_quantitative_experiment_rord} and~\ref{fig:more_quantitative_experiment_defacto}, the images are sourced from the RORD-Val and the DEFACTO~\cite{coco} splicing section, respectively. These results demonstrate that SmartEraser effectively removes target objects and consistently outperforms existing approaches.

\begin{figure*}[t]
  \centering
   \includegraphics[width=1\linewidth]{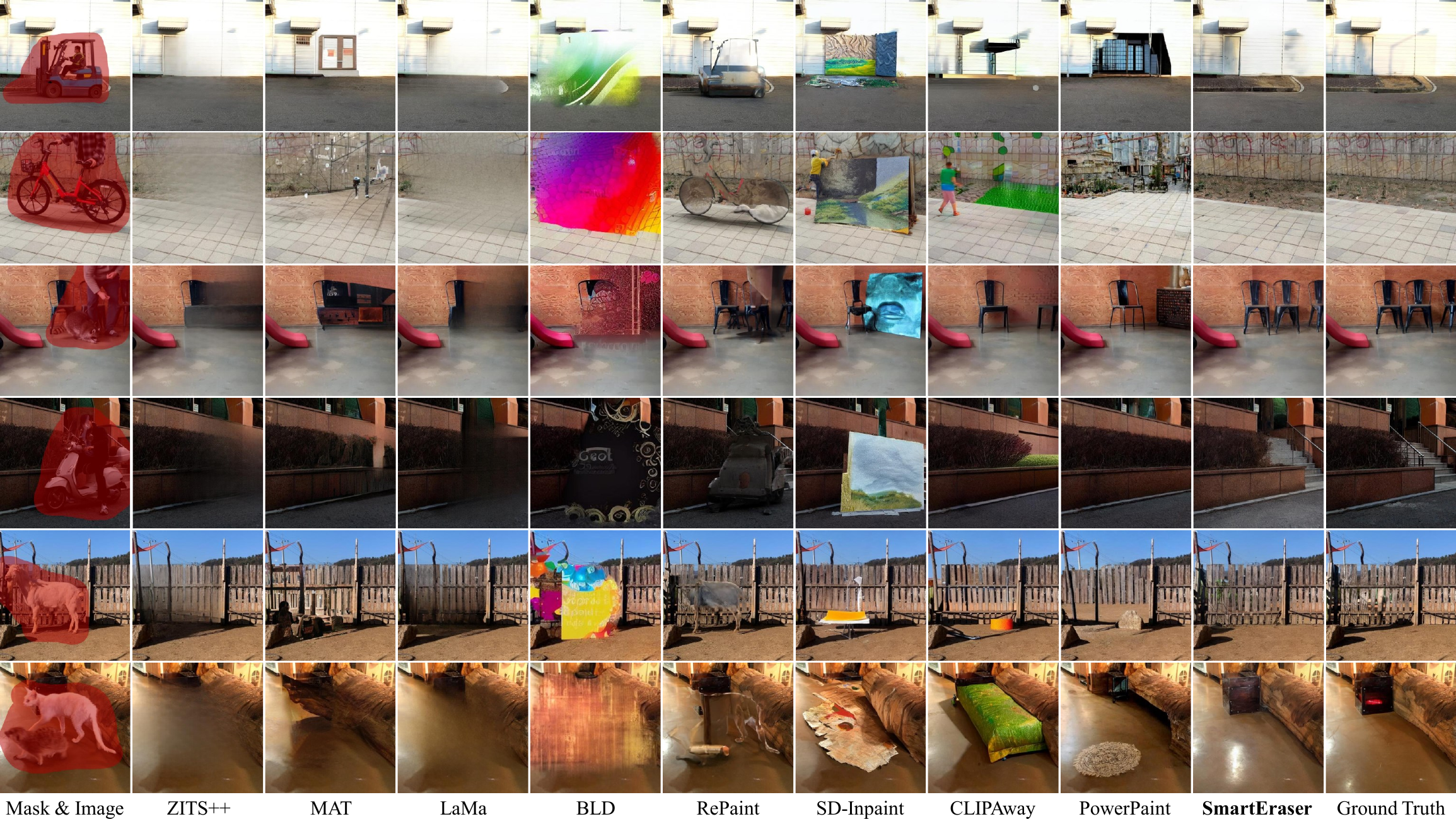}
   \vspace{-2em}
   \caption{Qualitative comparison of different methods in real-world user cases. The samples are sourced from RORD-Val.}
   \vspace{-0.5em}
   \label{fig:more_real_usercase_rord}
\end{figure*}

\begin{figure*}[t]
  \centering
   \includegraphics[width=1\linewidth]{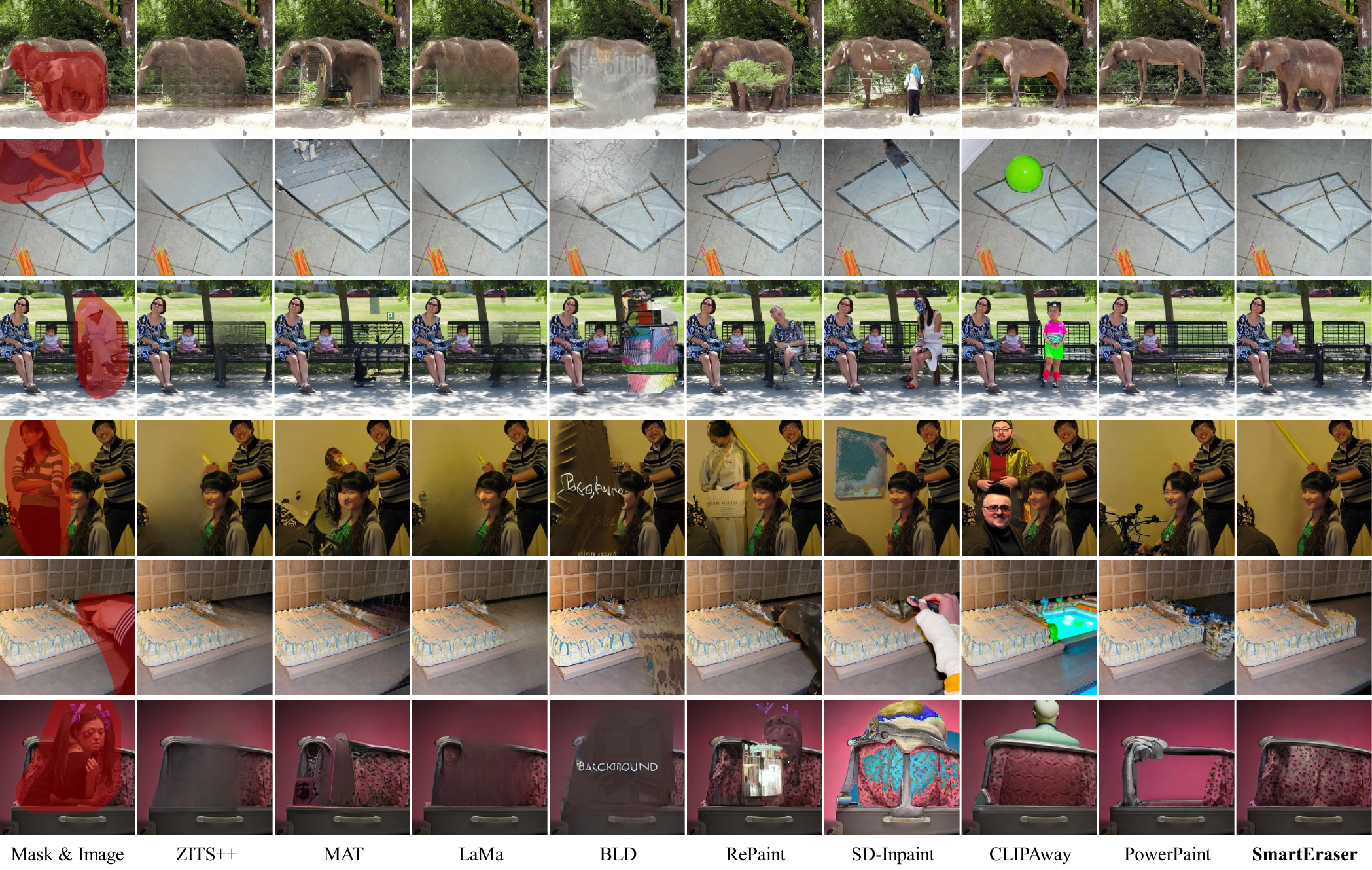}
   \vspace{-2em}
   \caption{Qualitative comparison of different methods in real-world user cases. The samples are sourced from the validation set in MSCOCO. Note that there are no ground truths.}
   \vspace{-0.5em}
   \label{fig:more_real_usercase_coco}
\end{figure*}

\noindent\textbf{More Real-world User Cases.}
To further demonstrate the ability of SmartEraser to smartly remove the target objects while preserving the surrounding background context, we present more real-world user cases compared with previous methods in Figures~\ref{fig:more_real_usercase_rord} and~\ref{fig:more_real_usercase_coco}. The images are sourced from the RORD-Val and the MSCOCO~\cite{coco} validation set. These examples show the capability of SmartEraser to accurately remove objects and maintain high consistency in background context within masks when facing diverse and complex real-world user cases.

\noindent\textbf{Erasing Scene Text.}
As shown in Figure~\ref{fig:text_erase}, we explore the capability of SmartEraser when applying it to erasing scene text. We observe that based on the proposed novel paradigm, our SmartEraser can remove scene text seamlessly without hurting the surrounding context.

\noindent\textbf{Partial object removal.} 
We conducted experiments on erasing partial objects. The results in the following Figure~\ref{fig:incomplete_object_sup} show that SmartEraser effectively identifies and removes the partial object specified by user mask.

\noindent\textbf{Occluded object removal.}
The following Figure~\ref{fig:overlapping_object_sup} shows that SmartEraser effectively removes occluded objects while successfully inpainting background. During Syn4Removal synthesis, objects are placed based on IoU criteria (Formula~3 in paper), allowing some occlusions.





\begin{figure}[t]
  \centering
   \includegraphics[width=1\linewidth]{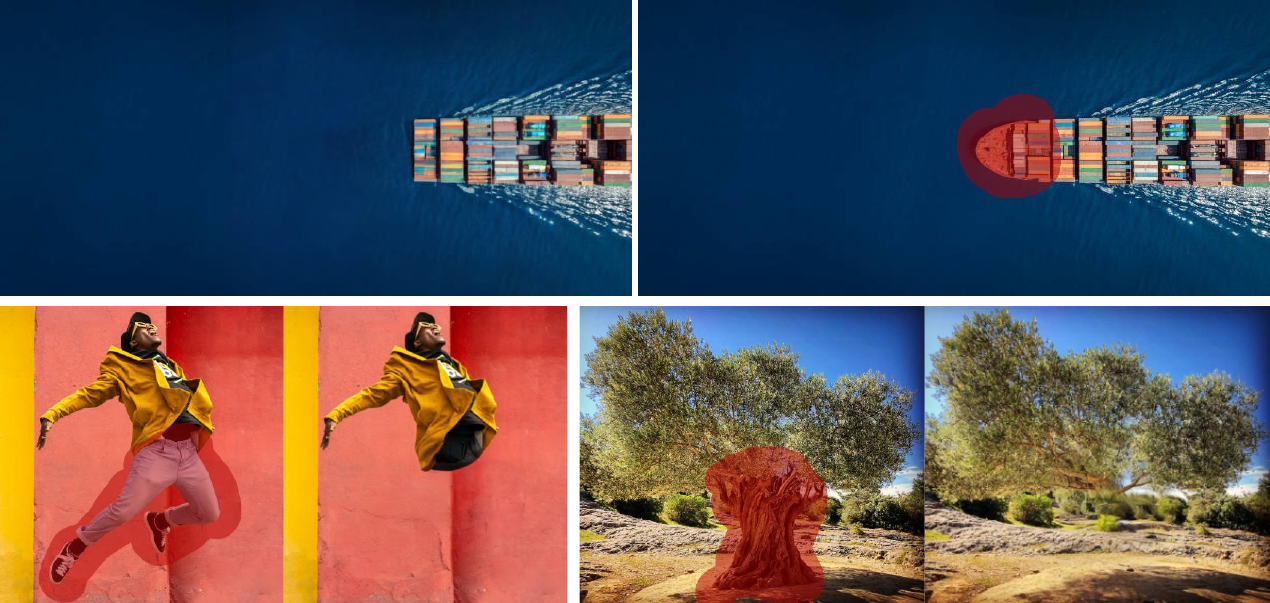}
   \vspace{-2em}
   \caption{Qualitative results for partial object removal.}
   \vspace{-1em}
   \label{fig:incomplete_object_sup}
\end{figure}

\begin{figure}[t]
  \centering
   \includegraphics[width=1\linewidth]{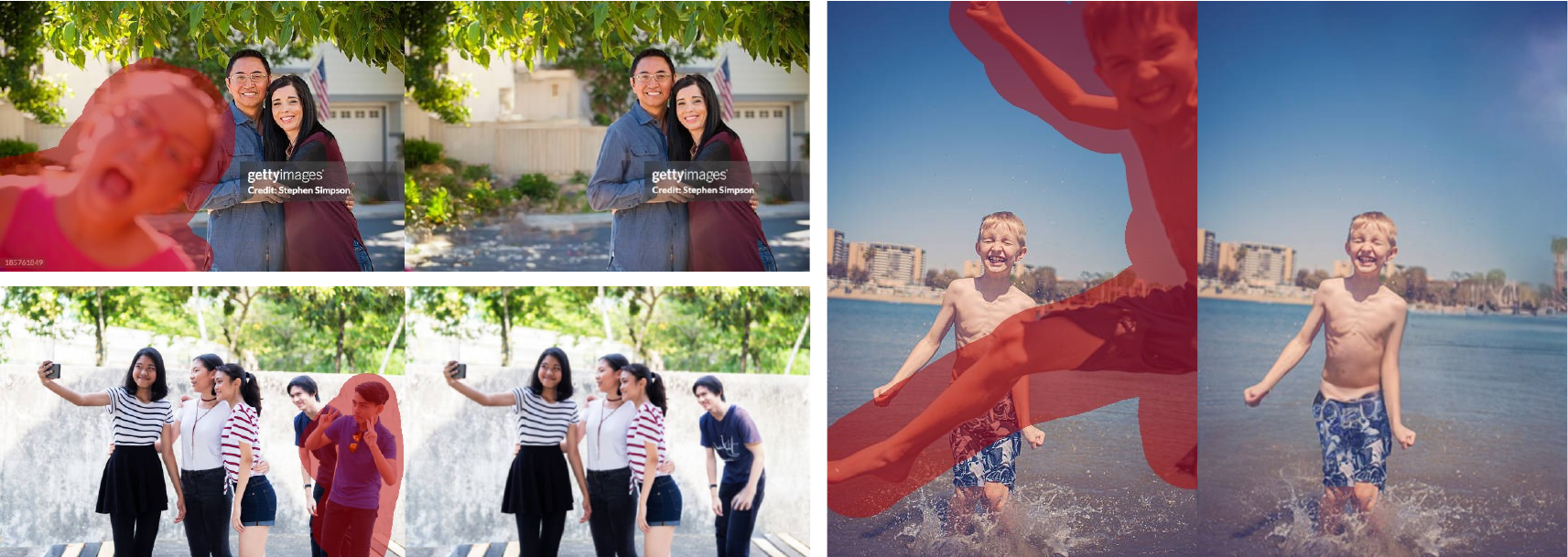}
   \vspace{-2em}
   \caption{Qualitative results for occluded object removal.}
   \vspace{-1em}
   \label{fig:overlapping_object_sup}
\end{figure}

\begin{figure}[t]
  \centering
   \includegraphics[width=1\linewidth]{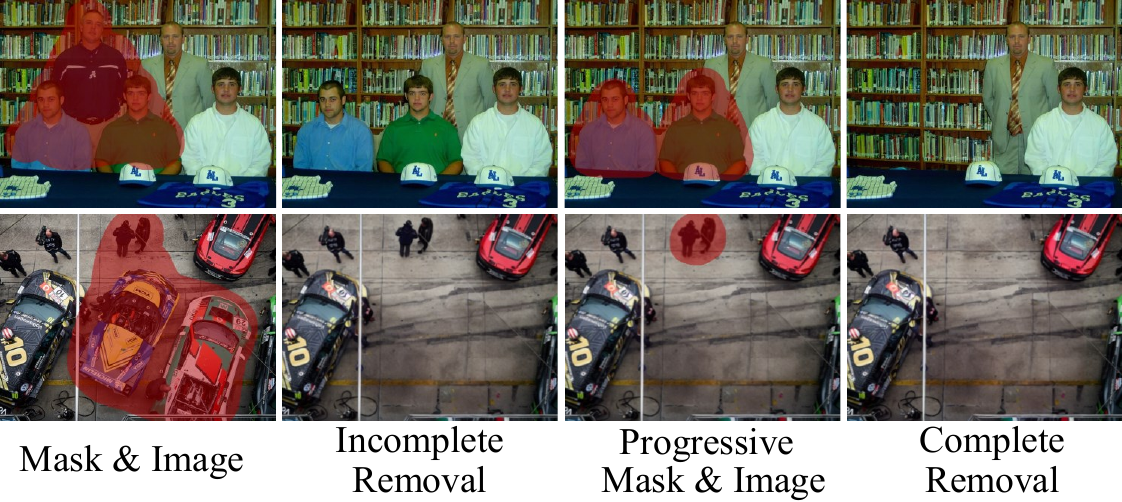}
   \vspace{-2em}
   \caption{Progressive object removal when needing to remove too many objects.}
   \vspace{-1em}
   \label{fig:limitation_progressive}
\end{figure}

\section{Limitations}

Although the strong performance is shown in the aforementioned Tables and Figures, we recognize there is a potential limitation. Our novel masked-region guidance helps the model identify the target objects and avoid unintentional regeneration, but if the masked region contains too many objects,
the model may fail to completely remove objects. This is because our paradigm may fail to identify all the target objects in the masks while considering some of them as background context.
A straightforward solution is to draw masks progressively to remove all that wants to be removed. The results of progressive removal are shown in Figure~\ref{fig:limitation_progressive}. We observe that when facing multiple objects, our SmartEraser can progressively remove all target objects.

\clearpage

{
    \small
    \bibliographystyle{ieeenat_fullname}
    \bibliography{main}
}

\end{document}